\documentclass[lettersize,journal]{IEEEtran}
\usepackage{amsmath}
\usepackage{array}
\usepackage[caption=false,font=normalsize,labelfont=sf,textfont=sf]{subfig}
\usepackage{textcomp}
\usepackage{stfloats}
\usepackage{url}
\usepackage{verbatim}
\usepackage{graphicx}
\usepackage{cite}
\usepackage{amssymb}
\usepackage{dsfont}
\usepackage{hyperref}
\usepackage{booktabs}  
\usepackage{multirow}
\usepackage{xcolor}
\usepackage[flushleft]{threeparttable}
\usepackage[ruled,vlined,linesnumbered]{algorithm2e}
\usepackage{algorithmic}

\definecolor{myred}{RGB}{255,0,0}
\definecolor{mybrown}{RGB}{188,121,118}
\definecolor{myblue}{RGB}{72,116,203}

\begin{document}

\title{OGScene3D: Incremental Open-Vocabulary 3D Gaussian Scene Graph Mapping for Scene Understanding}

\author{Siting Zhu, Ziyun Lu, Guangming Wang, Chenguang Huang, Yongbo Chen, I-Ming Chen~\IEEEmembership{Fellow,~IEEE}, Wolfram Burgard~\IEEEmembership{Fellow,~IEEE}, Hesheng Wang~\IEEEmembership{Senior Member,~IEEE}
\thanks{\quad Corresponding Author: Hesheng Wang (email: wanghesheng@sjtu.edu.cn)}
\thanks{\quad Siting Zhu, Ziyun Lu, and Yongbo Chen are with the Shanghai Jiao Tong University, China. Guangming Wang is with the University of Cambridge, England. Chenguang Huang and Wolfram Burgard are with the University of Technology Nuremberg, Germany.
I-Ming Chen is with the Nanyang Technological University, Singapore.
Hesheng Wang is with Department of Automation, Key Laboratory of System Control and Information Processing of Ministry of Education, State Key Laboratory of Avionics Integration and Aviation System-of-Systems Synthesis, Shanghai Key Laboratory of Navigation and Location Based Services, Shanghai Jiao Tong University, Shanghai, China.}
}



\maketitle

\begin{abstract}
    Open-vocabulary scene understanding is crucial for robotic applications, enabling robots to comprehend complex 3D environmental contexts and supporting various downstream tasks such as navigation and manipulation. However, existing methods require pre-built complete 3D semantic maps to construct scene graphs for scene understanding, which limits their applicability in robotic scenarios where environments are explored incrementally. To address this challenge, we propose OGScene3D, an open-vocabulary scene understanding system that achieves accurate 3D semantic mapping and scene graph construction incrementally. Our system employs a confidence-based Gaussian semantic representation that jointly models semantic predictions and their reliability, enabling robust scene modeling. Building on this representation, we introduce a hierarchical 3D semantic optimization strategy that achieves semantic consistency through local correspondence establishment and global refinement, thereby constructing globally consistent semantic maps. Moreover, we design a long-term global optimization method that leverages temporal memory of historical observations to enhance semantic predictions. By integrating 2D-3D semantic consistency with Gaussian rendering contribution, this method continuously refines the semantic understanding of the entire scene.Furthermore, we develop a progressive graph construction approach that dynamically creates and updates both nodes and semantic relationships, allowing continuous updating of the 3D scene graphs. Extensive experiments on widely used datasets and real-world scenes demonstrate the effectiveness of our OGScene3D on open-vocabulary scene understanding. Codes will be available at \href{https://github.com/IRMVLab/OGScene3D}{https://github.com/IRMVLab/OGScene3D}.
\end{abstract}


\begin{figure}
    \centering
    \includegraphics[width=1\linewidth]{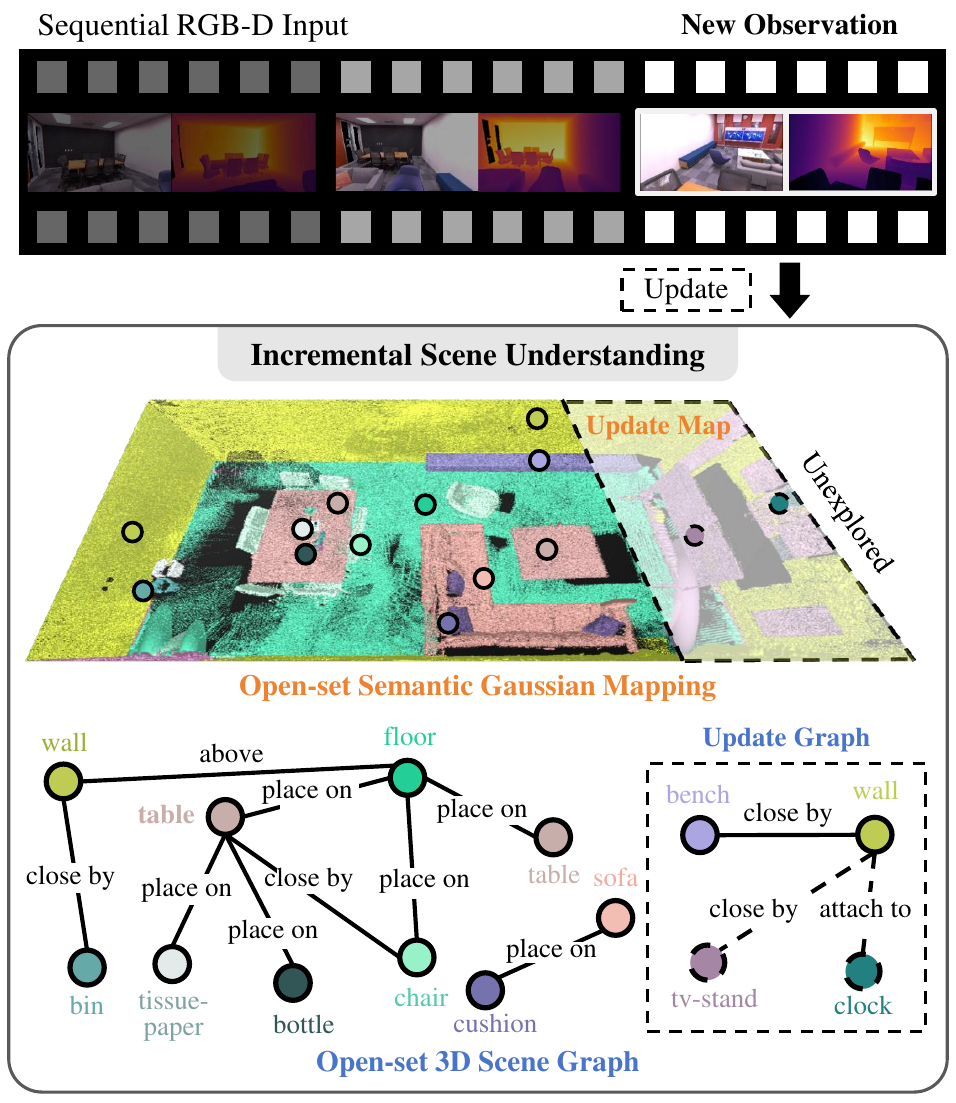}
    \caption{Our OGScene3D achieves incremental open-vocabulary scene understanding by continuously updating scene graphs and semantic Gaussian maps.
    }
    \label{fig:first-page}
\end{figure}

\section{Introduction}
\IEEEPARstart{3}{D} scene understanding refers to the process of perceiving and interpreting 3D environment from sensor input, aiming to comprehend both semantic properties of entities and their spatial relationships within the scene. It is the cornerstone of robotic applications~\cite{gu2024conceptgraphs, werby23hovsg} and computer vision~\cite{kim20193, wu2023incremental}.
However, most previous works~\cite{wu2021scenegraphfusion, hughes2022hydra, gay2019visual} are limited to closed-set scene understanding as they rely on predefined semantic categories. This limitation poses significant challenges  for robotic applications in real-world environments, where robots frequently encounter novel objects during manipulation and navigation tasks. Therefore, open-vocabulary scene understanding is critical for downstream tasks.

Recent works~\cite{gu2024conceptgraphs, werby23hovsg, conceptfusion, deng2024openobj, yang2025opengs, zhai2025panogs} leverage the capabilities of foundation models in visual perception, semantic reasoning, and relational understanding to achieve open-vocabulary scene understanding. Although these approaches, including traditional representation-based ~\cite{gu2024conceptgraphs, werby23hovsg, conceptfusion}, Neural radiance field (NeRF)~\cite{mildenhall2021nerf}-based methods~\cite{deng2024openobj}, and 3D Gaussian Splatting (3DGS)~\cite{kerbl20233d}-based methods~\cite{yang2025opengs, zhai2025panogs}, have demonstrated their ability to perform open-vocabulary semantic mapping and scene graph construction, they still exhibit several limitations.

First, current scene representations face significant challenges in the fundamental trade-offs between semantic expressiveness, semantic precision,  and memory efficiency. 
Traditional representation-based~\cite{gu2024conceptgraphs, werby23hovsg, conceptfusion, RAZER} approaches represent each semantic region in 3D space with a single feature vector extracted from vision-language model (VLM).
However, these feature vectors cannot directly output semantic meanings and instead require similarity computation against predefined text descriptions to assign specific semantic categories to regions.
NeRF-based methods~\cite{deng2024openobj}, despite achieving high-fidelity reconstruction, struggle with unbounded scenes and slow rendering speeds.
Among 3DGS-based semantic mapping approaches, methods that integrate semantic feature with each Gaussian~\cite{zhai2025panogs} provide rich semantic representation but significantly increase memory requirements. In contrast, methods~\cite{yang2025opengs} that assign discrete label to each Gaussian are memory-efficient but highly susceptible to errors in 2D semantic segmentation, as inaccuracies from individual frames can propagate throughout the reconstruction. 
Second, current open-vocabulary semantic mapping approaches perform semantic updates in a frame-by-frame manner, where each new observation only influences the semantic representation of currently visible regions. This local processing lacks global optimization capabilities, resulting in semantic inconsistencies across the scene and inadequate representation in sparsely observed areas.
Third, existing works can only perform offline graph construction based on prebuilt 3D semantic maps. Such offline approach is impractical for robotic applications where environments are explored incrementally and tasks need to be performed during progressive scene exploration based on scene understanding.

To address these challenges, we propose OGScene3D, an open-vocabulary scene understanding system that achieves precise 3D semantic mapping and scene graph construction incrementally. 
To achieve effective trade-offs between semantic expressiveness, precision, and memory efficiency, we employ a confidence-based 3DGS semantic representation that incorporates only semantic label and confidence into each Gaussian. 
Through iterative refinement of semantic confidence and label, our method explicitly models semantic uncertainty arising from segmentation ambiguities and geometric errors, effectively filtering unreliable predictions to  enhance semantic precision.
Additionally, we generate explicit semantic descriptions for each identified region through multi-view visual analysis and VLM-based captioning during progressive scene understanding.
Moreover, we introduce hierarchical 3D semantic optimization that integrates local frame-to-scene associations with global semantic refinement, enabling semantic consistency across the entire scene. 
In addition, we maintain a memory buffer that preserves historical semantic and geometric information throughout the mapping process. Based on this buffer, we employ long-term 3D optimization that mitigates catastrophic forgetting and enhances semantic representation across the entire scene, especially in areas with limited visibility. 
Furthermore, to enable online scene understanding, we introduce a progressive graph construction approach that employs confidence-based selective updates to dynamically create and refine both semantic nodes and relationships. This approach produces incrementally-built 3D scene graph that expands with environmental exploration.

Our contributions can be summarized as follows: 
\begin{itemize}
    \item We propose a photorealistic open-vocabulary scene understanding system, which achieves accurate 3D semantic mapping and scene graph construction progressively based on 3D Gaussian representation. We employ a novel confidence-based Gaussian semantic representation for accurate scene modeling.
    \item  We introduce hierarchical 3D semantic optimization that integrates local frame-to-scene associations with global semantic refinement for consistent 3D semantic mapping. We also design a long-term global 3D optimization strategy that leverages long-term historical information to enhance semantic representation.
    \item  We employ a progressive scene graph construction approach that dynamically establishes and updates both nodes and semantic relationships, enabling continuous optimization of the 3D scene graphs. 
    \item  Extensive evaluations are conducted on three challenging datasets, Replica~\cite{straub2019replica}, ScanNet~\cite{dai2017scannet}, 3RScan~\cite{Wald2019RIO}, and real-world scenes, to demonstrate our method achieves superior performance compared with existing open-vocabulary scene understanding methods.
\end{itemize}
\section{Related Work}
\subsection{3D Representation and Semantic Mapping}
Traditional semantic mapping employs explicit 3D representation, such as points~\cite{tateno2017cnn, li2016incremental}, surfels~\cite{mccormac2017semanticfusion, tateno2015real, runz2018maskfusion, runz2017co}, voxel~\cite{wang2023semlaps}, Truncated Signed Distance Fields (TSDF)~\cite{schmid2022panoptic, mccormac2018fusion++, grinvald2019volumetric, zheng2019active, xu2019mid} and mesh~\cite{tian2022kimera, rosinol2021kimera, hughes2022hydra, rosinol2020kimera, miao2024volumetric, rosinol20203d} for dense mapping. 
SemanticFusion~\cite{mccormac2017semanticfusion} keeps track of the class probability distribution for each surfel and updates those probabilities based on the predictions of Convolutional Neural Networks (CNN). Panoptic Multi-TSDFs~\cite{schmid2022panoptic} employs multi-resolution volumetric map representation to capture long-term object-level scene changes. Kimera~\cite{rosinol2020kimera} utilizes bundled raycasting to build an accurate global 3D mesh and semantically annotate the mesh.

However, these representations suffer from limited geometric resolution and fail to achieve high-fidelity 3D reconstruction. 
Recently, Neural radiance field (NeRF)~\cite{mildenhall2021nerf} and 3D Gaussian Splatting (3DGS)~\cite{kerbl20233d} have shown promising capability in dense reconstruction and novel view synthesis. Building upon these scene representations, various semantic mapping methods have been developed, including NeRF-based semantic mapping~\cite{zhi2021place, zhi2022ilabel, li2024dns, zhai2024nis, zhu2024sni, zhu2026sni} and 3DGS-based semantic mapping~\cite{li2024sgs, zhu2024semgauss, ji2024neds, li2024hi, li2024gs3lam}. 
Semantic-NeRF~\cite{zhi2021place} extends NeRF to jointly encode semantics with appearance and geometry.
SNI-SLAM~\cite{zhu2024sni} introduces coarse-to-fine semantic implicit representation for precisely constructing semantic maps.
SemGauss-SLAM~\cite{zhu2024semgauss} incorporates semantic feature embedding into 3D Gaussian representation for dense semantic mapping.
While these methods achieve incremental dense semantic mapping, they rely on predefined semantic categories, limiting them to closed-set semantic mapping. However, in real-world robotic applications, robots often encounter novel object types that were not included in the training data, which highlights the need for open-set semantic mapping.

Open-set semantic mapping typically leverages foundation models to achieve zero-shot perception. Recognize Anything Model (RAM)~\cite{zhang2024recognize} enables recognition of any common categories, while models like SAM~\cite{kirillov2023segment}, MobileSAM~\cite{zhang2023faster}, and FastSAM~\cite{zhao2023fast} achieve image segmentation. Moreover, YOLO-World~\cite{cheng2024yolo} and Grounding DINO~\cite{liu2024grounding} perform zero-shot object detection. Recent works leverage the above foundation models for open-set semantic reconstruction, including traditional representation-based~\cite{gu2024conceptgraphs, werby23hovsg, conceptfusion, RAZER}, NeRF-based~\cite{deng2024openobj, kerr2023lerf, liu2023weakly, liao2024ov, yuan2024uni}, 3DGS-based methods~\cite{yang2025opengs, martins2024ovo, zhou2024feature, ye2024gaussian, zhai2025panogs, wu2024opengaussian, qin2024langsplat, shi2024language, bhalgat2024n2f2, deng2025omnimap}.
ConceptGraphs~\cite{gu2024conceptgraphs} extracts a single CLIP~\cite{radford2021learning} feature for each SAM-segmented object from 2D RGB-D images. These object-level features are then projected into 3D space and are processed through multi-view association and fusion to achieve open-set semantic mapping.
HOV-SG~\cite{werby23hovsg} employs feature clustering of zero-shot embeddings for open-set 3D semantic segmentation using CLIP~\cite{radford2021learning} and SAM~\cite{kirillov2023segment}. 
OmniMap~\cite{deng2025omnimap} uses YOLO-World~\cite{cheng2024yolo} for object detection and extracts semantic features from detection results for open-set mapping, but is limited by incomplete detection coverage.
OpenGS-SLAM~\cite{yang2025opengs} incorporates semantic label into each Gaussian for semantic representation and performs confidence-based 2D label consensus to ensure consistent labeling across multiple views for mapping.
However, these works suffer from imprecise semantic boundaries and cannot directly output semantic meanings due to representing semantic regions with single feature vectors.  Additionally, many existing methods are susceptible to errors in 2D semantic segmentation that propagate into the 3D reconstruction.
In this paper, we propose a confidence-based 3DGS semantic representation that effectively models semantic uncertainty for 3D semantic mapping. Moreover, we introduce hierarchical 3D semantic optimization that performs frame-to-scene associations and global semantic optimization for accurate mapping. 

\subsection{3D Scene Graphs}
While semantic mapping focuses on high-level semantic understanding of scene objects, the spatial relationships between objects are captured through 3D scene graph construction. 3D scene graphs provide a compact and efficient representation of scenes using graph structures, where nodes represent objects and edges encode inter-object relationships.
Traditional scene graph prediction methods~\cite{chen2019knowledge, suhail2021energy, tang2020unbiased, tang2019learning, woo2018linknet, xu2017scene, yang2018graph,zellers2018neural} primarily focused on inferring object relationships directly from 2D images. Subsequent approaches~\cite{armeni20193d, wald2020learning, wang2023vl} infer scene relationships directly from 3D scene representations. 3DSSG~\cite{wald2020learning} employs Graph Convolutional Network (GCN)~\cite{kipf2016semi} to generate a semantic scene graph from a 3D point cloud. 
Building upon this, several works~\cite{wu2021scenegraphfusion, wu2023incremental, hughes2022hydra, kim20193, gay2019visual} enable dynamically constructing graphs and object relationships as the scene is progressively reconstructed. 
SceneGraphFusion~\cite{wu2021scenegraphfusion} introduces an attention-based method that can predict scene graph from partial and incomplete 3D data for incremental graph construction.

However, these methods are constrained by closed-vocabulary setting, which limits their generalization ability as they cannot recognize relationships involving novel objects. Recent works~\cite{wang2025gaussiangraph, gu2024conceptgraphs, conceptfusion, deng2024opengraph, koch2024open3dsg, chen2024clip} utilize large language models (LLM)~\cite{achiam2023gpt} and large VLMs~\cite{liu2023visual} to achieve open-vocabulary scene graph construction. 
ConceptGraphs~\cite{gu2024conceptgraphs} estimates the spatial relationships of objects after the entire image sequence has been processed, and instructs the LLM to describe the likely spatial relationship.
Nevertheless, these methods are limited to offline graph construction based on prebuilt 3D semantic map.
Such offline approaches present significant limitations for robotic applications, where environments are dynamically explored and robots need to execute tasks during progressive scene exploration.
In this paper, we propose progressive graph construction, which enables dynamic node addition, relationship establishment, and confidence-based node updates for incremental open-set graph construction as the scene is explored.

\begin{figure*}
    \centering
    \includegraphics[width=\linewidth]{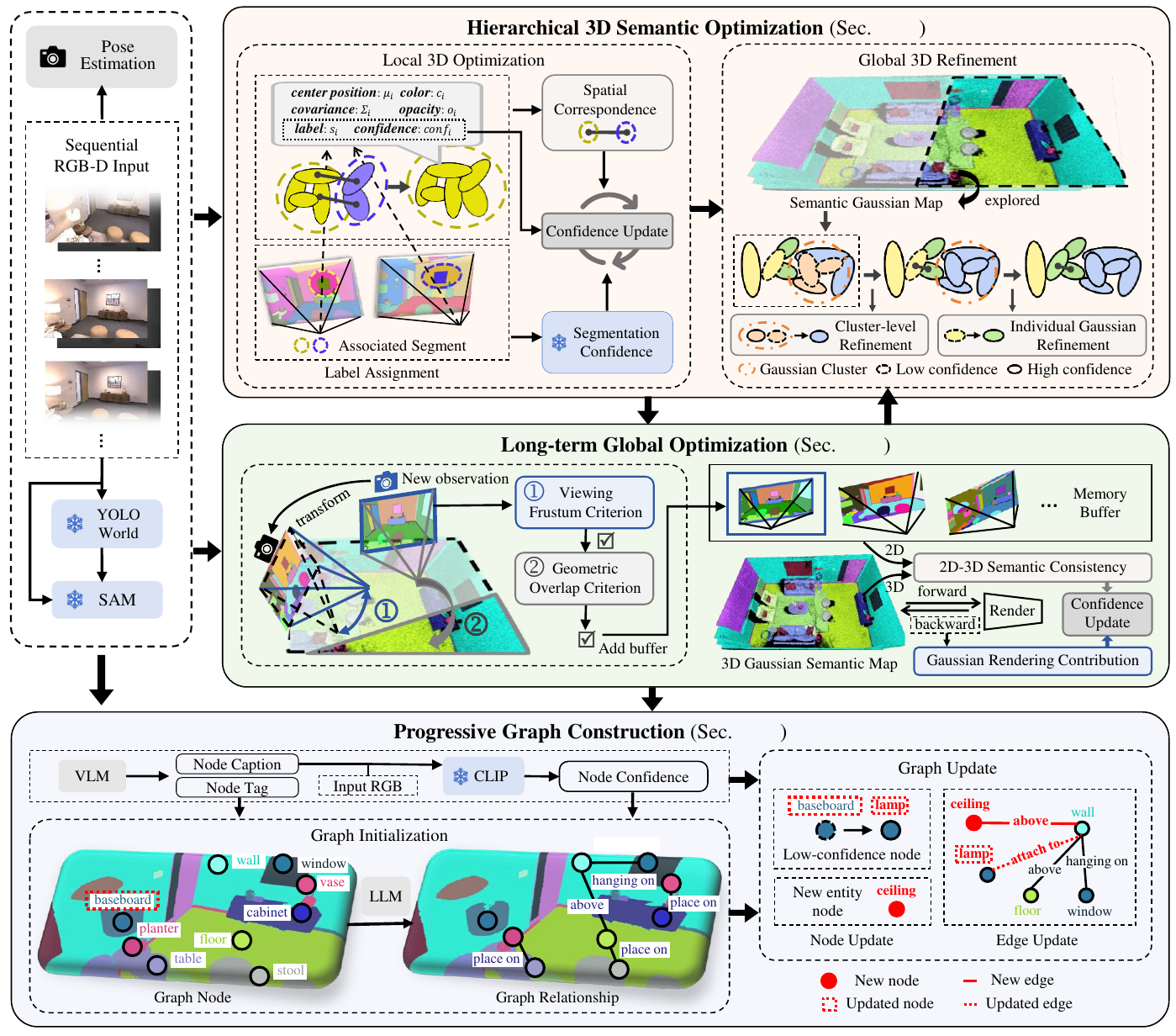}
    \put(-131.5, 438.7){\small{\ref{sec:Hierarchical 3D Semantic Optimization}}}
    \put(-146.5, 255.2){\small{\ref{sec:Long-term global optimization}}}
    \put(-193.4, 129.0){\small{\ref{sec:Progressive Graph Construction}}}
    \caption{\textbf{An overview of OGScene3D.} Given sequential RGB-D images, our method achieves open-vocabulary semantic mapping through hierarchical 3D semantic optimization and long-term global optimization. At the same time, our method enables incremental scene understanding through progressive scene construction. 
    }
    \label{fig:overview}
\end{figure*}

\section{Method}
The overview of our method is shown in Fig.~\ref{fig:overview}. Given an input RGB-D frames $I = \{C_t, D_t\}_{t=1}^{N}$, where $C_t$ and $D_t$ represent the RGB image and depth map respectively, we perform open-vocabulary scene understanding of the environment incrementally.
Sec.~\ref{sec:Confidence-based scene representation} presents the confidence-based scene representation of our system. 
Sec.~\ref{sec:Hierarchical 3D Semantic Optimization} introduces hierarchical 3D semantic optimization, achieving globally consistent semantic mapping through local optimization and global refinement.
Sec.~\ref{sec:Long-term global optimization} describes our long-term global optimization strategy, which leverages historical information to enhance semantic representation.
Sec.~\ref{sec:Progressive Graph Construction} introduces progressive graph construction that incrementally updates both captions of entities (nodes) and inter-entity relationships (edges), enabling comprehensive scene understanding.

\subsection{Confidence-based Scene Representation}
Existing open-vocabulary scene understanding methods primarily employ features or labels for semantic representation. Feature-based approaches~\cite{gu2024conceptgraphs, werby23hovsg} extract high-dimensional feature vectors from CLIP~\cite{radford2021learning} and associate them with 3D regions for semantic modeling. However, these feature vectors cannot directly output explicit semantic descriptions and struggle with interpretability. Label-based approach~\cite{yang2025opengs} assigns discrete semantic labels directly to individual 3D primitives,  but lacks robustness when handling inconsistent segmentation results across multiple frames.
To address these limitations, we introduce confidence-based scene representation that associates both individual 3D primitives and semantic entities with confidence. This modeling enables our system to achieve robustness against segmentation inconsistencies and reconstruction errors through confidence-aware updates.

\label{sec:Confidence-based scene representation}
\noindent\textbf{Semantic 3D Gaussian Representation.}\hspace*{5pt}
We employ a set of Gaussian primitives $\mathcal{G}$ for 3D scene modeling. Each 3D Gaussian $G_i$ is composed of its center position $\mu_{i}=\{x_{i}, y_{i}, z_{i}\}$, covariance $\Sigma_{i}\in \mathbb{R}^{3\times 3}$, opacity $o_{i}$, color $c_{i}$, semantic label $s_{i}$, and semantic confidence $conf_{i}$.
To save memory storage, $G_i$ stores a semantic label $s_i$ instead of high-dimensional feature vectors for semantic representation.
$\mathcal{G}$ is represented by:
\begin{equation}
\mathcal{G}(\boldsymbol{x}) = e^{-\frac{1}{2}(x-\mu)^T {\Sigma}^{-1} (x-\mu)}, 
\end{equation}
where $\Sigma$ is parameterized as $\Sigma=R S {S}^T {R}^T$ for optimization. $R$ represents 3D rotation and $S$ is a scaling matrix.

\noindent\textbf{3D Scene Graph Structure.}\hspace*{5pt}
Building upon the semantic 3D Gaussian representation, we construct a scene graph, denoted $\mathcal{S}=\{\mathcal{N}, \mathcal{E}\}$, to explicitly model the higher-level structure and semantic relationships within the scene. The nodes $\mathcal{N}$ are formed by grouping Gaussian primitives that share the same semantic label. Each node $n_i \in \mathcal{N}$ represents a distinct semantic entity and is associated with a node confidence $\mathcal{C}_i$. The edges $\mathcal{E}$ capture the spatial relationships between these nodes, where $e_{ij} \in \mathcal{E}$ denotes the relationship between node $n_i$ and $n_j$.

\noindent\textbf{Color and Depth Rendering.}\hspace*{5pt}
Following the differentiable rasterization~\cite{max2002optical}, rendered pixel color $C_p$ and depth $D_p$ can be calculated by blending Gaussians sorted in depth order: 
\begin{equation}
\begin{aligned}
C_p =  \sum_{i \in N} c_{i} \alpha_{i} \prod_{j=1}^{i-1} (1 - \alpha_j), \\
D_p =  \sum_{i \in N} d_{i} \alpha_{i} \prod_{j=1}^{i-1} (1 - \alpha_j),
\end{aligned}
\end{equation}
where $\alpha_{i}=o_i\mathcal{G'}_i$, $G_i'$ is 2D projection of $G_i$.

\noindent\textbf{Confidence-based Semantic Rendering.}\hspace*{5pt}
Unlike continuous color and depth values, semantic labels are discrete categorical values that cannot be directly blended. 
Therefore, for 2D semantic label rendering, we employ a dual-weighted voting approach that integrates both geometric reliability and semantic uncertainties. Specifically, each Gaussian votes for its semantic class, weighted by its accumulated alpha value (representing geometric opacity) and semantic confidence. The final semantic label $L_p$ at pixel $p$ is determined as the class $k$ with the highest accumulated weighted votes:
\begin{equation}
L_p = \operatorname*{argmax}_k ( \sum_{i \in N}  \vec{s}_i \cdot conf_i \cdot \alpha_{i} \prod_{j=1}^{i-1} (1 - conf_j \cdot \alpha_j) ), 
\end{equation}
where $\vec{s}_i \in \mathbb{R}^K$ represents the one-hot encoded semantic vector for the $i$-th Gaussian, $K$ denotes the total number of semantic classes. Note that the initialization and update of the per-Gaussian confidence are detailed in Sec.~\ref{sec:Hierarchical 3D Semantic Optimization} and Sec.~\ref{sec:Long-term global optimization}.
This formulation ensures that only Gaussians with both high geometric reliability (opacity) and high semantic confidence can significantly influence the final semantic label assignment.

\noindent\textbf{Pose Estimation.}\hspace*{5pt} 
We utilize DROID-SLAM~\cite{teed2021droid} for online pose estimation from the RGB-D stream. As camera poses are incrementally computed, we simultaneously perform semantic mapping and scene graph construction, which enables our OGScene3D to be a fully online system.

\subsection{Hierarchical 3D Semantic Optimization}
\label{sec:Hierarchical 3D Semantic Optimization}
For open-vocabulary semantic mapping, a fundamental challenge lies in maintaining semantic consistency across multiple viewpoints. While foundation models like SAM~\cite{kirillov2023segment} excel at segmenting individual frames, they lack the capability to ensure consistent semantic labeling for the same objects across different views, thereby limiting coherent 3D semantic reconstruction.
To address this challenge, we propose hierarchical 3D semantic optimization that operates at two levels: local optimization establishes cross-frame correspondences through geometric and semantic constraints, followed by global refinement to ensure semantic consistency across the entire 3D scene. 
Our approach begins with open-set segmentation to obtain accurate 2D semantic masks, then applies the hierarchical optimization to achieve consistent 3D semantic mapping.

\subsubsection{Open-set Segmentation}
We perform open-set 2D segmentation using foundation models for subsequent 3D semantic reconstruction. Specifically, we utilize SAM~\cite{kirillov2023segment} $Seg(\cdot)$ to generate primary segmentation $M_s^t=Seg(C_t)$ using uniform grid points as prompts, where this model groups similar pixels into potential semantic regions. Meanwhile, to enhance segmentation accuracy of scene objects, we also employ YOLO-World~\cite{cheng2024yolo} to detect objects and obtain their associated labels. These detections then serve as prompts for SAM~\cite{kirillov2023segment} to produce more refined segmentation $M_d^t$. While $M_d^t$ demonstrates higher precision for detected objects, it may miss segments for objects that the detector fails to identify. Therefore, we fuse $M_s^t$ with $M_d^t$ to obtain the final segmentation $M_f^t$ for frame $C_t$. $M_f^t$ includes all segments from $M_d^t$, supplemented by segments from $M_s^t$ that have a pixel overlap ratio below threshold $\tau_m$ with corresponding segment in $M_d^t$, ensuring comprehensive scene coverage while maintaining object-level precision. 

\begin{figure}
    \centering
    \includegraphics[width=\linewidth]{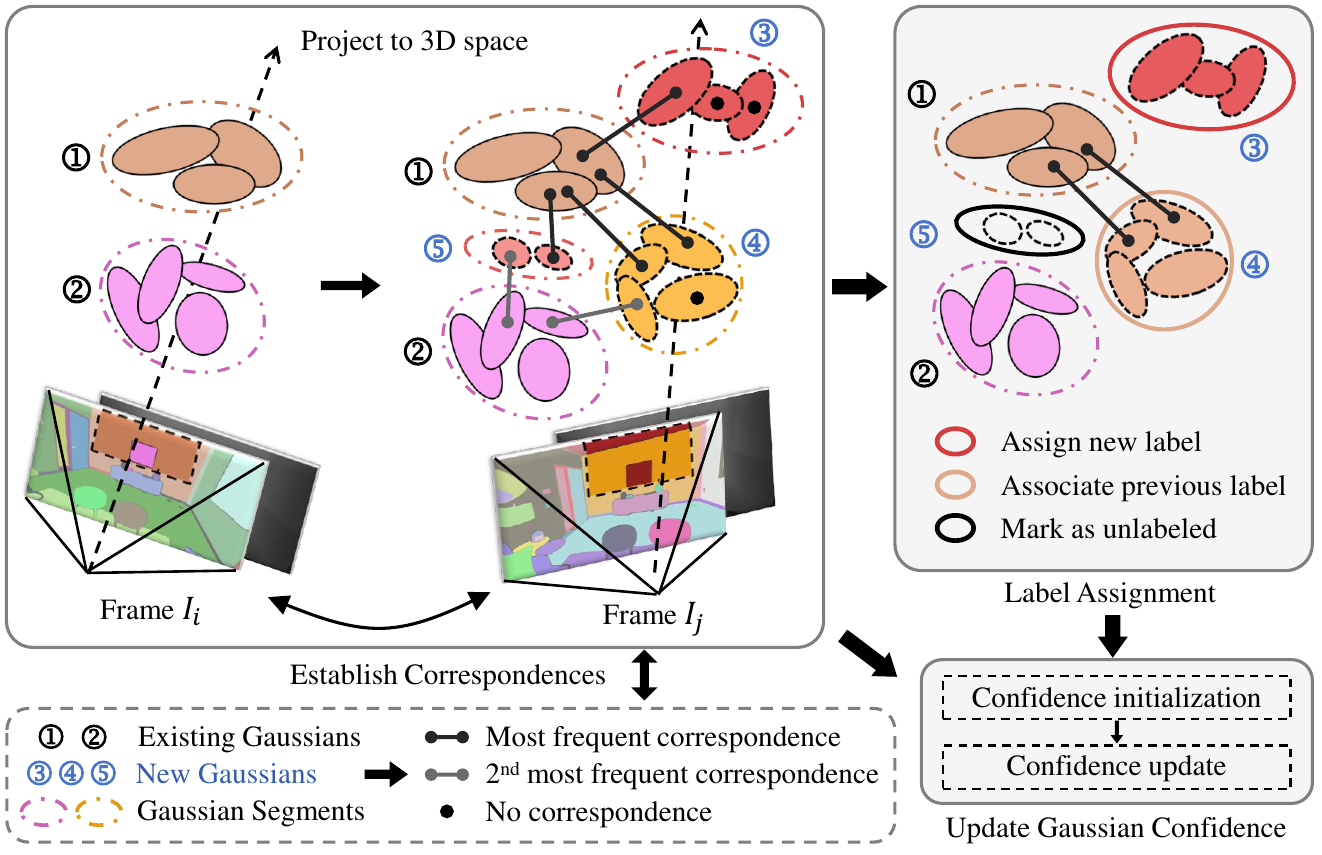}
    \caption{\textbf{Local 3D Optimization.} When a new keyframe $I_j$ arrives, new 3D Gaussians are initialized from its segmented regions using depth information. To maintain semantic consistency with the existing scene representation, KNN correspondence is established between new Gaussian segments and existing scene Gaussians. Each new Gaussian segment is then evaluated using semantic ambiguity ratio $r(m)$ and spatial-semantic coherence $p(m)$ to determine label assignment. Finally, Gaussian semantic confidence is updated using spatial consistency measure and semantic confidence.
    }
    \label{fig:local3d}
\end{figure}

\subsubsection{Local 3D Optimization}
\label{Local 3D optimization}
The 2D segmentation $M_f^t$ obtained above only maintains semantic consistency within individual frames. In other words, $M_f^t$ exhibits inconsistent semantic labels for the same objects across different viewpoints, making them unsuitable for direct 3D semantic reconstruction.
To address this cross-view inconsistency, we introduce local 3D optimization that first establishes spatial-semantic associations between Gaussian segments, then dynamically updates semantic confidence to mitigate segmentation and correspondence uncertainties.

Given an RGB-D input, we first initialize semantic 3D Gaussians using 2D segmentation $M_f^1$ and depth map $D_1$ from the first frame $I_1$, which serves as the initial keyframe. For each semantic region in $M_f^1$, we project the corresponding pixels into 3D space using $D_1$ and camera intrinsics to obtain 3D points. These points serve as initialization centers for Gaussians, where Gaussians initialized from the same semantic region are assigned the same semantic label. As new keyframe $I_t$ arrives, we utilize its segments $m \in M_f^t$ and depth $D_t$ to initialize new Gaussians $\{\mathcal{G}_m\}_{m \in M_f^t}$.
However, these newly initialized Gaussians are assigned semantic labels from $M_f^t$, which may not be consistent with the semantic labels in the existing scene representation. 

\noindent\textbf{Spatial-semantic Association.}\hspace*{5pt}
To maintain semantic consistency, we introduce segment-level spatial-semantic association. This approach establishes robust correspondences between new and existing Gaussian segments, addressing the unreliability of individual associations that arises from pose estimation and reconstruction uncertainties, as shown in Fig.~\ref{fig:local3d}.

We utilize k-nearest neighbor (KNN) search to establish individual correspondences between 3D Gaussians in new segments $\{\mathcal{G}_m\}_{m \in M_f^t}$ and existing scene Gaussians, then aggregate these correspondences to enable segment-level associations. For each Gaussian in the new segments, we find its nearest neighbors based on spatial distance in 3D space. Based on these individual spatial correspondences, we define two key segment-level measures for each new Gaussian segment $\mathcal{G}_m$:
\begin{equation}
\label{eq4}
r(m) = \frac{c_2(m)}{c_1(m)}, \quad p(m) = \frac{|{\{G_i \in \mathcal{G}_m : \hat{l}_i = l_m^*}\}|}{|\mathcal{G}_m|},
\end{equation}
where $c_1(m)$ and $c_2(m)$ represent the counts of the most and second most frequent semantic labels across all KNN matches for Gaussians in $\mathcal{G}_m$; $p(m)$ denotes the proportion of Gaussians in $\mathcal{G}_m$ that correspond to the most frequent semantic label $l_m^*$, where $\hat{l}_i$ is the KNN-matched label for Gaussian $G_i$. 
$r(m)$ measures the semantic ambiguity ratio. A higher $r(m)$ suggests comparable frequencies between the top two labels, indicating ambiguous semantic classification, while a lower $r(m)$ indicates clear semantic dominance.
$p(m)$ represents the spatial-semantic coherence of the Gaussian segment.  Higher values indicate stronger spatial consistency between the new segment and existing labeled regions, suggesting reliable geometric-semantic correspondence.

Based on these measures, the semantic label reassignment follows three cases:
(i) $r(m) \geq \tau_\text{valid}$: all Gaussians in $\mathcal{G}_m$ are marked as unlabeled and excluded from semantic rendering. This criterion effectively filters out ambiguous segments caused by SAM failures in real-world scenarios, thereby preventing erroneous segmentation from degrading scene representation quality.
(ii) $r(m) < \tau_\text{valid}$ and $p(m) \geq \tau_p$: all Gaussians in $\mathcal{G}_m$ are assigned the semantic label $l$. This segment-level label association enhances robustness against potential pose estimation or reconstruction errors. 
(iii) $r(m)<\tau_\text{valid}$ and $p(m) < \tau_p$: all Gaussians in $\mathcal{G}_m$ are assigned a novel semantic label, as these segments likely represent newly observed objects due to their low spatial correspondence with existing labeled Gaussians.
The above labeling strategy enables our system to filter out ambiguous segmentation artifacts, maintain semantic coherence for previously observed objects, and dynamically incorporate new semantic classes as they appear in the scene.

While the segment-level association strategy provides robust label assignment, discrete semantic labels alone may not fully encode the inherent uncertainty of the segmentation and correspondence processes. To provide a more comprehensive semantic representation, we augment each Gaussian's semantic label with a confidence $conf_{i}$ that quantifies the reliability of the semantic assignment. This confidence-aware representation enables more accurate semantic updates during incremental mapping and better handling of uncertain regions.

\noindent\textbf{Confidence Initialization.}\hspace*{5pt}
For each Gaussian $G_i$, we first initialize its confidence as $conf_i = c_{sam}(i)$. $c_{sam}(i)$ denotes the semantic confidence obtained by back-projecting the 2D pixel-wise confidence from the SAM output $Seg(\cdot)$ onto the corresponding 3D Gaussians. Note that the SAM model outputs confidence values within a narrow range $[0.88, 1.0]$. To facilitate more effective confidence updates during incremental mapping, we linearly rescale these values to the range $[0.5, 1.0]$. This rescaling provides a wider dynamic range for confidence refinement, allowing for greater differentiation between highly confident and less confident semantic assignments as new observations are incorporated.

\noindent\textbf{Confidence Update.}\hspace*{5pt}
Then, during incremental mapping, we dynamically refine the confidence $conf_i$ for existing scene Gaussians by leveraging the correspondences established in the spatial-semantic association.
Specifically, for each matched scene Gaussian, we update its confidence by incorporating the spatial correspondence ratio and semantic consistency information from new viewpoints.
When a new keyframe $I_t$ arrives, each matched scene Gaussian $G_i$ is assigned a spatial consistency measure $p^t(i) = p(m)$ based on its correspondence with new segment $\mathcal{G}_m$. Here, $p(m)$ (Eq.~\ref{eq4}) quantifies how well the new segment spatially aligns with existing labeled regions, also indicating the correspondence reliability for Gaussian $G_i$. 
We then update the confidence of $G_i$ according to the semantic consistency between its current label and the corresponding semantic observation in the new keyframe:
(i) confidence increases when the label of $G_i$ matches its corresponding observation; (ii) confidence decreases when the labels differ. This confidence update is formulated as:
\begin{equation}
\label{eq5}
\small
conf_i' = \begin{cases}
conf_i + c_{sam}^t(i) \cdot p^t(i) \cdot (1-conf_i) & G_i \in \mathcal{C}_t \\
conf_i - c_{sam}^t(i) \cdot p^t(i) \cdot conf_i & G_i \notin \mathcal{C}_t,
\end{cases}
\end{equation}
where $conf_i'$ represents the updated confidence after incorporating observations from keyframe $I_t$, $c_{sam}^t(i)$ is the semantic confidence of the current keyframe $I_t$, and $\mathcal{C}_t$ represents the set of matched Gaussians whose semantic labels align with their corresponding observations in keyframe $I_t$. 
Note that confidence updates are not applied to scene Gaussians which correspond to unlabeled segments or segments with novel semantic labels. These Gaussians lack semantically consistent observations from novel views that are essential for reliable confidence refinement.
The above formulation allows us to dynamically adjust semantic label confidence by leveraging both geometric correspondence and semantic cues from multiple views, leading to increasingly reliable scene understanding as more observations are integrated.

\subsubsection{Global 3D Refinement}
\label{Global 3D refinement}
Through local association and optimization between sequential frames, we obtain a consistent semantic 3D Gaussian representation, where each semantic label is associated with a confidence indicating its reliability. However, incremental reconstruction from partial observations inevitably introduces accumulated errors in the semantic map. To refine the semantic representation, we conduct global refinement at regular intervals determined by cumulative camera motion. This process consists of cluster-level and individual Gaussian refinement, detailed in Algorithm \ref{algo:global_refinement}.

\begin{algorithm}[t]
\caption{Global 3D Refinement}
\label{algo:global_refinement}
\KwIn{3D semantic Gaussians $\mathcal{G} = \{G_i\}$ where $G_i = \{\mu_i, s_i, conf_i\}$; \\
\quad\quad\quad Semantic clusters $\mathcal{C} = \{C_k\}$ where $C_k \subset \mathcal{G}$;
\textbf{Thresholds:} $\tau_\text{low}$, $\tau_\text{high}$, search radius $r_\text{search}$;
}
\KwOut{Refined 3D semantic Gaussians $\mathcal{G}'$}

\BlankLine
\textbf{Cluster-level Refinement:}\\
Initialize set of processed Gaussians $\mathcal{P} \gets \emptyset$ \\
\ForEach{semantic cluster $C_k$}{
    Compute average confidence $A_k = \frac{1}{|C_k|}\sum_{G_i \in C_k} conf_i$
    
    \If{$A_k < \tau_\text{low}$}{
        Initialize label collection $L_k \gets \emptyset$ \\
        \ForEach{Gaussian $G_i \in C_k$}{
            Find $G_j$: nearest Gaussian outside $C_k$ with $conf_j > \tau_\text{high}$\\
            $L_k \gets L_k \cup \{s_j\}$
        }
        $s_\text{new} \gets \arg\max_{s} |\{s' \in L_k : s' = s\}|$ \\
        \ForEach{Gaussian $G_i \in C_k$}{
            $s_i \gets s_\text{new}$ \\
            $\mathcal{P} \gets \mathcal{P} \cup \{G_i\}$ 
        }
    }
}

\BlankLine
\textbf{Individual Gaussian Refinement:}\\
\ForEach{Gaussian $G_i \in \mathcal{G} \setminus \mathcal{P}$}{
    \If{$conf_i < \tau_\text{low}$}{
        $\mathcal{N}_i \gets$ Find Gaussians within $r_\text{search}$ of $\mu_i$ where $conf > \tau_\text{high}$\\
        
        \If{$\mathcal{N}_i \neq \emptyset$}{
            $G_j \gets \arg\min_{G_k \in \mathcal{N}_i} \|\mu_i - \mu_k\|$\\
            $s_i \gets s_j$\\
            $conf_i \gets 0.5$
        }
    }
}

\Return{$\mathcal{G}'$}
\end{algorithm}

\noindent\textbf{Cluster-level Refinement.}\hspace*{5pt}
First, we employ semantic refinement on 3D semantic clusters. Let $C_k$ represents a cluster containing spatially adjacent Gaussians that share the same semantic label. For each cluster $C_k$, we evaluate its label reliability by computing the average confidence $A_k$ of all Gaussians within the cluster. When $A_k<\tau_\text{low}$, we consider this cluster to be potentially mislabeled and discard its original semantic label for refinement. 
Its new semantic label is determined by performing KNN search for each Gaussian in $C_k$ to find its nearest high-confidence Gaussian ($conf_{i} > \tau_\text{high}$) outside $C_k$ in 3D space. The cluster is then assigned the most frequent semantic label among all these neighboring Gaussians.
This cluster-level refinement strategy effectively reduces semantic noise while preserving the geometric structure of scene representation, as it leverages both confidence-based filtering and spatial consistency for label refinement.
The confidence-based filtering enables identifying unreliable labels of Gaussian clusters, while spatial consistency ensures that the refined labels are coherent with the surrounding well-labeled regions through proximity-based label assignment.
Notably, the refinement process only updates the semantic labels while preserving the original confidence of individual Gaussians, ensuring that the uncertainty information from the local 3D optimization is retained.
We record all Gaussians that have been refined during cluster-level processing to exclude them from subsequent individual-level refinement, avoiding redundant modifications.

\noindent\textbf{Individual Gaussian Refinement.}\hspace*{5pt}
Then, we refine the remaining individual Gaussians with low confidence, which typically appear at object boundaries or in regions with uncertain semantic segmentation. For each low-confidence Gaussian $G_i = \{\mu_i, s_i,\}$ where $conf_i < \tau_\text{low}$, we consider its semantic label $s_i$ unreliable and perform label refinement. The refinement process is conducted by spatial nearest-neighbor search within a radius $r_\text{search}$ around $\mu_i$. We identify candidate Gaussians whose confidence exceeds a threshold $\tau_\text{high}$, then select the spatially closest candidate $G_j = \{\mu_j, s_j\}$ based on the Euclidean distance $|\mu_i - \mu_j|$. 
The refined Gaussian becomes $G_i' = \{\mu_i, s_j\}$, effectively preserving its spatial position while adopting the semantic label from the nearby high-confidence Gaussian. Additionally, we update the confidence to an intermediate value $conf_i' = 0.5$ to prevent the refined label from being immediately overwritten in subsequent global optimization when the Gaussian is temporarily unobserved. 
Such Gaussian-based refinement maintains local geometric details while effectively propagating reliable semantic information from high-confidence regions, leading to more coherent semantic segmentation results.

\subsubsection{Loss Functions}
The above processes focus on semantic optimization. For optimization of Gaussian geometric and appearance parameters, we construct depth loss and color loss by comparing rendered color and depth with input RGB-D values. SSIM term~\cite{kerbl20233d} is added to color loss:
\begin{equation}
\begin{gathered}
 \mathcal{L}_{c} = (1-\lambda)\|C(\mathcal{G},T_w^{i}) - C_i\| + \lambda \mathcal{L}_\text{SSIM}, \\
 \mathcal{L}_{d} = \|D(\mathcal{G},T_w^{i}) - D_i\|,
\end{gathered}
\end{equation}
where $\lambda=0.2$, $T_w^{i}$ is camera pose of frame $i$ in world coordinate, $C(\mathcal{G},T_w^{i})$ and $D(\mathcal{G},T_w^{i})$ denote the rendered color and depth. The overall loss function is the weighted sum of the above losses: 
\begin{equation}
 \mathcal{L} = \lambda_{c}\mathcal{L}_{c} + \lambda_{d}\mathcal{L}_{d},
 \end{equation}
where $\lambda_{c}$, $\lambda_{d}$ are weighting coefficients.

\subsection{Long-term Global Optimization}
\label{sec:Long-term global optimization}
While the previous optimization stages effectively maintain local consistency (Sec.~\ref{Local 3D optimization}) and perform global refinement (Sec.~\ref{Global 3D refinement}), these stages rely on sufficient observations to obtain accurate semantic labels and confidences. However, for regions with sparse observations, their semantic predictions remain uncertain due to insufficient viewpoints and temporal information. To address the semantic uncertainty in under-observed areas, we introduce a long-term global optimization approach that leverages historical information to enhance semantic predictions. 
Our approach maintains a memory buffer that stores observations collected at different times, which cover complementary spatial regions of the scene.
Based on this buffer, we update semantic confidence through 2D-3D semantic consistency and Gaussian rendering contribution, improving semantic understanding of the entire scene.
Note that our long-term optimization only updates confidence rather than semantic labels. As this optimization operates on previously mapped regions with established semantic assignments, modifying labels directly could compromise the consistency of existing representations. Instead, updating only confidence maintains the stability of scene representations while identifying unreliable regions for subsequent global refinement.

\subsubsection{Memory Buffer Construction}
We maintain a memory buffer $\mathcal{M}=\{M_i\}_{i \in N}$ storing representative frames and their corresponding viewpoints, which are selected based on spatial coverage and distinctiveness. When a new keyframe $I_t=\{C_t,D_t\}$ arrives, we project pixels to 3D space using $D_t$ to obtain point clouds $\textbf{p}_t$.  We then employ a view-geometry dual evaluation strategy to determine the inclusion of $I_t$ in $\mathcal{M}$. 

\noindent\textbf{Viewing Frustum Criterion.}\hspace*{5pt}
The first criterion quantifies view distinctiveness by calculating the percentage of points in $\textbf{p}_t$ that fall within the viewing frustum of the most recent memory buffer $M_r$:
\begin{equation}
\begin{gathered}
\small
R_\text{in} = \frac{|\{p^\prime(x,y,z) \in T_{I_t}^{M_r} \textbf{p}_t : |\frac{x}{z}|  \leq \tan(\frac{f_h}{2}) \land |\frac{y}{z}| \leq \tan(\frac{f_v}{2})\}|}{|\textbf{p}_t|}, \\
f_h = 2 \cdot \arctan(\frac{W}{2f}) \quad 
f_v = 2 \cdot \arctan(\frac{H}{2f}), \\
\end{gathered}
\end{equation}
where $T_{I_t}^{M_r}$ represents the rigid transformation from camera frame $I_t$ to $M_r$, $W$ and $H$ are image width and height, $f$ is known focal length, $p^\prime(x,y,z)$ represents transformed 3D points, and $\land$ denotes the logical conjunction operation. 
This initial criterion, $R_\text{in} < \tau_\text{long}$, verifies that sufficient viewpoint disparity exists between $I_t$ and $M_r$, serving as a preliminary geometric validation. 
However, viewpoint disparity alone does not guarantee observation of new scene geometry, as the current frame might observe previously captured regions from different viewpoints.

\noindent\textbf{Geometric Overlap Criterion.}\hspace*{5pt}
Therefore, we further evaluate spatial overlap by establishing correspondence between $\textbf{p}_t$ and point clouds $\textbf{p}(\mathcal{M})$ derived from the entire memory buffer $\mathcal{M}$:
\begin{equation}
R_\text{overlap} = \frac{|\{p \in \textbf{p}_t : \exists q \in \textbf{p}(\mathcal{M}), |p-q| < r_\text{search}\}|}{|\textbf{p}_t|},
\end{equation}
where $r_\text{search}$ is the spatial search radius. The geometric overlap criterion, $R_\text{overlap} < \tau_\text{long}$, ensures that the candidate frame $I_t$ contributes substantial additional geometric coverage to the memory buffer. 

If both criteria are satisfied, the keyframe $I_t$ and its corresponding viewpoint are incorporated into the memory buffer $\mathcal{M}$. This selection provides a compact yet geometrically representative memory buffer for long-term optimization.
Based on $\mathcal{M}$, we update semantic confidence of the 3D Gaussian map through 2D-3D semantic consistency and gaussian rendering contribution. We utilize multiple viewpoints $\{v_i\}_{i \in N_\text{view}}$ from the buffer to refine semantic predictions.

\subsubsection{2D-3D Semantic Consistency}
We leverage the consistency between 2D segmentation and 3D semantic map as a reliability indicator to update semantic predictions. Regions exhibiting high semantic consistency are more likely to be correctly labeled, while inconsistent regions may suffer from segmentation ambiguities or mapping errors.
Therefore, we employ this consistency to refine the semantic confidence of our Gaussian map.
To quantify this consistency, we establish semantic correspondence between 2D segmentation results from SAM and our global 3D map. We first project the 3D semantic map onto the camera viewpoint $v_i$ to generate 2D rendered semantic $S^{v_i}$. We then compute the Intersection over Union (IoU) between $S^{v_i}$ and the 2D segmentation to quantify the 2D-3D semantic alignment.
For each semantic class $k$, we calculate a class-wise confidence $P_\text{2D-3D}^k$ across the entire image domain $\Omega$:
\begin{equation}
    P_\text{2D-3D}^k = \frac{\sum_{p \in \Omega} \mathbb{I}[
    Seg(M_{v_i})'_p \wedge (S^{v_i}_p=k)]}{\sum_{p \in \Omega} \mathbb{I}[
    Seg(M_{v_i})'_p \vee (S^{v_i}_p=k)]},
\end{equation}
where $M_{v_i} \in \mathcal{M}$, $\mathbb{I}[\cdot]$ is the indicator function, $\vee$ is the logical OR operator, , $Seg(M_{v_i})'_p$ represents the remapped SAM segmentation mask at pixel $p$.  This remapping aligns the original segment labels with the scene semantic categories while preserving the original segmentation boundaries.
To obtain pixel-wise confidence $P_\text{2D-3D}(p)$, we assign each pixel with the class-wise confidence $P_\text{2D-3D}^k$ of its corresponding semantic class $k$:
\begin{equation}
    P_\text{2D-3D}(p) = P_\text{2D-3D}^k, \quad \text{if } S^{v_i}_p = k.
\end{equation}
This pixel-wise confidence $P_\text{2D-3D}(p)$ quantifies the semantic consistency between single-view segmentation and the global 3D semantic map at each spatial location. To utilize $P_\text{2D-3D}(p)$ for updating 3D Gaussian representation, we establish a mapping from 2D pixels to existing 3D Gaussians in the scene. For each pixel $p$ with associated depth from memory buffer $M_{v_i}$, we project it into the 3D space to obtain corresponding 3D points. We then associate each point with its nearest Gaussians through KNN matching. 
Through this pixel-to-Gaussian association, we can assign the pixel-wise confidence $P_\text{2D-3D}(p)$ to their corresponding 3D Gaussians,  obtaining Gaussian-wise confidence update factor $P_\text{2D-3D}(i)$ for each Gaussian $i$ in the scene representation.

\subsubsection{Gaussian Rendering Contribution}
Building upon the consistency measure $P_\text{2D-3D}$, we further enhance the semantic confidence update by incorporating the varying influence of individual Gaussians on the rendered image. 
Intuitively, Gaussians that contribute more significantly to visible pixels encode richer information and should therefore have proportionally larger impact on their semantic confidence updates.
To quantify this contribution, we propose a gradient-based method that leverages the differentiable nature of the Gaussian rendering process.
Specifically, we first formulate the rendered value $R_p$ at pixel $p$ as a weighted sum of semantic contributions from all Gaussians that influence this pixel:
\begin{equation}
R_p =  \sum_{i \in N}  \vec{s}_i \cdot conf_i \cdot \alpha_{i} \prod_{j=1}^{i-1} (1 - conf_j \cdot \alpha_j), 
\end{equation}
where $N$ represents the set of contributing Gaussians, $\vec{s}_i \in \mathbb{R}^K$ is the one-hot semantic vector for the $i$-th Gaussian, $K$ denotes the total number of semantic classes. 
To achieve effective gradient-based update, we formulate two complementary loss functions that capture different aspects of the rendered confidence distribution:
\begin{equation}
L_{sq} = \frac{1}{2}\sum_{p \in P} R_p^2, \quad
L_{lin} = \sum_{p \in P} R_p.
\end{equation}
The squared loss $L_{sq}$ amplifies the impact of pixels with larger rendered values, making Gaussians that contribute to high-confidence regions have substantially larger gradients. The linear loss $L_{lin}$ treats all pixels equally, providing a linear baseline measure of overall confidence.
Then, by computing the gradients of these losses with respect to the semantic vectors $\vec{s}_i$, we can quantify the impact of each Gaussian on the rendered image: 
\begin{equation}
\frac{\partial L_{sq}}{\partial \vec{s}_i} = \sum_{p \in P} R_p \cdot \frac{\partial R_p}{\partial \vec{s}_i},  \quad
\frac{\partial L_{lin}}{\partial \vec{s}_i} = \sum_{p \in P} \frac{\partial R_p}{\partial \vec{s}_i}  
\end{equation}
The ratio of these gradients forms our confidence update factor $P_\text{conf}$, which adaptively weights the influence of each Gaussian on the semantic confidence update:
\begin{equation}
P_\text{conf}(i) = \frac{\frac{\partial L_{sq}}{\partial \vec{s}_i}}{\frac{\partial L_{lin}}{\partial \vec{s}_i} } = \frac{\sum_{p \in P} R_p \cdot \frac{\partial R_p}{\partial \vec{s}_i}}{\sum_{p \in P} \frac{\partial R_p}{\partial \vec{s}_i} + \epsilon}.
\end{equation}

\subsubsection{Long-term Confidence Update}
To integrate long-term information into the confidence update process, we project the buffer frames $\{\mathcal{M}_{v_i}\}$ into 3D space and associate them with the current scene representation. For Gaussians whose semantic labels are consistent with observations from historical frames, we increase their confidence through a positive update term. Conversely, for Gaussians exhibiting label inconsistencies, we apply a negative update to reduce their confidence. This adaptive update strategy is formulated as:
\begin{equation}
    conf_i' = \begin{cases}
    conf_i + P_\text{2D-3D}(i) \cdot P_\text{conf}(i) \cdot (1-conf_i) \\
    conf_i - P_\text{2D-3D}(i) \cdot (1 - P_\text{conf}(i)) \cdot conf_i
    \end{cases}
\end{equation}
This formulation ensures that confidence is updated based on both spatial consistency ($P_\text{2D-3D}$) and rendering contribution ($P_\text{conf}$), while maintaining a normalized confidence range. By leveraging temporally distant observations, we improve semantic understanding in regions that may have been under-observed in recent frames, creating a more complete and consistent 3D semantic representation.

\begin{figure*}
    \centering
    \includegraphics[width=\linewidth]{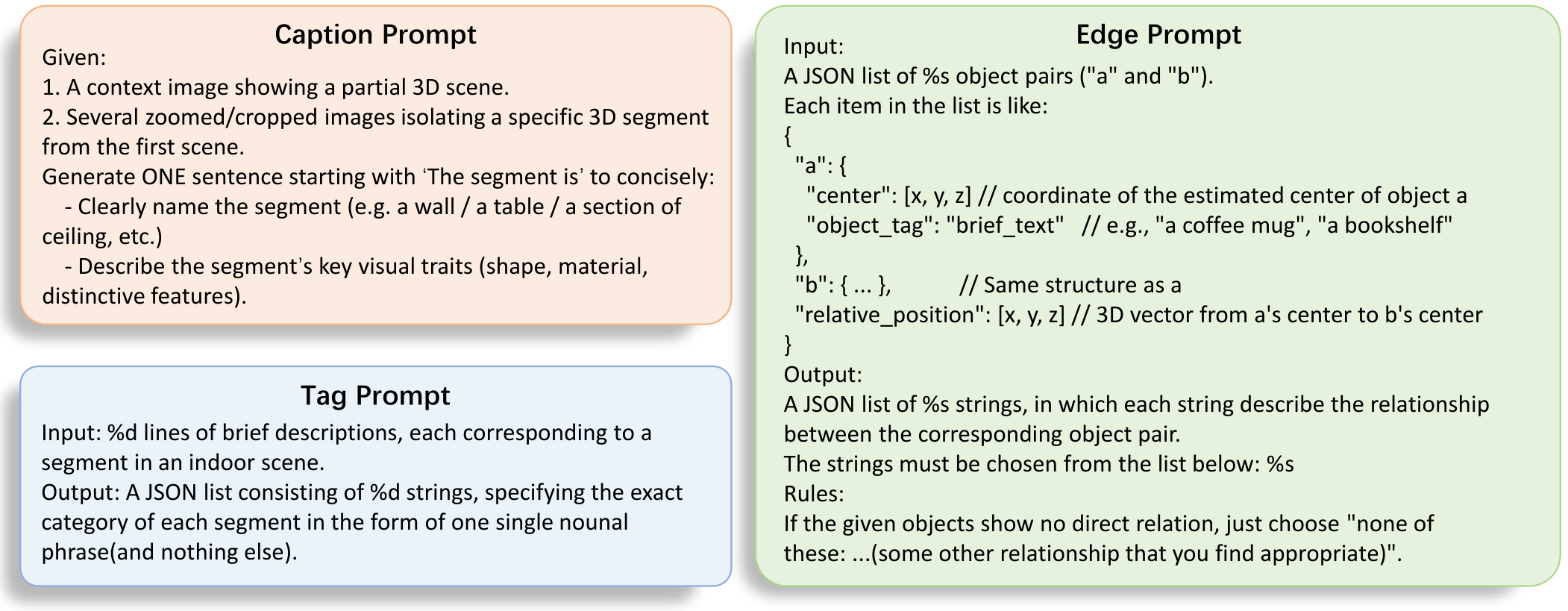}
    \caption{Designed prompts for progressive graph construction, including caption prompt $\mathcal{P}_\text{caption}$, tag prompt $\mathcal{P}_\text{tag}$, and edge prompt $\mathcal{P}_\text{edge}$. 
    }
    \label{fig:prompt}
\end{figure*}

\subsection{Progressive Graph Construction}
\label{sec:Progressive Graph Construction}
As described above, our incremental open-vocabulary semantic mapping approach partitions the environment into distinct semantic regions, with globally consistent label assignments across the mapped scene. However, these labels from SAM segmentation only serve as region identifiers, and fail to provide semantic meaning of the entities such as recognizing a region as a table or chair. 
To achieve a more comprehensive scene understanding and capture inter-entity relationships, we construct a 3D scene graph $\mathcal{S}=\{\mathcal{N}, \mathcal{E}\}$ during the incremental mapping process. 
Each node $n_i \in \mathcal{N}$ corresponds to a semantic region, which represents semantic entities such as furniture or structural elements like walls and ceiling.
Additionally, $n_i$ is associated with a natural language caption that describes its identity and attributes. The edge $e_{ij} \in \mathcal{E}$ encodes relationships between these entities. Our scene graph construction proceeds in two main phases: graph initialization, and graph update.

\subsubsection{Graph Initialization}
Scene graph initialization requires multiple observations from keyframes to establish accurate entity representations and reliable spatial relationships between entities. We utilize the first 12 keyframes for this initialization process. The process includes node captioning, node confidence initialization, and edge establishment. 

\noindent\textbf{Node Captioning.}\hspace*{5pt}
For each semantic entity identified across these keyframes, we extract two types of complementary visual information, including entity-centered crop with surrounding context, and masked images that preserve only the target entity. To balance information richness with computational efficiency, we select the largest crop and three largest masked images from different viewpoints to ensure comprehensive multi-view representation of the entity $i$:
\begin{equation}
\label{eq:oi}
    O_i = \{c_i, m_i^1, m_i^2, m_i^3\},
\end{equation}
where $c_i$ represents entity-centered crop, and $m_i^j$ denotes masked image from the $j$-th viewpoint.
We then leverage GPT-4o~\cite{achiam2023gpt} as a VLM to generate comprehensive entity captions: 
\begin{equation}
\label{eq:o_caption}
    \mathcal{O}_\text{caption}^i = VLM(O_i, \mathcal{P}_\text{caption}),
\end{equation}
where $\mathcal{P}_\text{caption}$ is caption prompt (Fig.~\ref{fig:prompt}). By presenting observations from different views, we enable the VLM to reason about the entity's complete 3D characteristics rather than relying on single-view perception. The masked images guide the VLM to focus on the target entity, while the contextual crop provides environmental cues that help determine the entity's role and function in the overall scene. This multi-view approach significantly enhances the quality and accuracy of the generated entity captions, which serve as the natural language descriptions associated with each node $n_i$ in our scene graph.

\noindent\textbf{Node Confidence Initialization.}\hspace*{5pt}
After obtaining the caption, we compute a node confidence $\mathcal{C}_i$ using CLIP~\cite{radford2021learning} to evaluate the semantic alignment between the visual representation and the generated caption:
\begin{equation}
\label{eq19}
\mathcal{C}_i = CLIP(c_i, \mathcal{O}_\text{caption}^i). 
\end{equation}
This confidence serves as a quality indicator for the generated captions and guides the subsequent update strategy. 
Note that during initialization, some entities may appear in fewer than three keyframes due to occlusion or limited field of view. In such cases, we still initialize these nodes with available visual information and continue to update their captions as more observations become available in subsequent frames.

\noindent\textbf{Edge Establishment.}\hspace*{5pt}
For edge establishment, we assess potential associations between entities in the scene by analyzing their spatial proximity. Through pairwise analysis between all entities, we determine meaningful spatial relationships and establish edges between the corresponding nodes in the scene graph.
To begin with, we define the set of points in entity $i$ that are proximate to entity $j$:
\begin{equation}
\label{eq:pij}
P_{i \to j} = \{p_i \in \mathbf{p}_i  : \exists p_j \in \mathbf{p}_j, |p_i - p_j| < \theta_e\},
\end{equation}
where $\mathbf{p}_i$ and $\mathbf{p}_j$ represent the point cloud clusters of entities $i$ and $j$, $\theta_e$ is the distance threshold for identifying proximate points between clusters.
We then calculate proximity interaction ratios as a measure of the spatial relationship:
\begin{equation}
\label{eq:rij}
R_{ij} = \max(\frac{|P_{i \to j}|}{|\mathbf{p}_i|}, \frac{|P_{j \to i}|}{|\mathbf{p}_j|}).
\end{equation}
Relationship will be established between entities $i$ and $j$ when $R_{ij} > \tau_e$, forming a set of spatially adjacent entity pairs $\mathcal{R} = \{(i,j) \mid R_{ij} > \tau_e\}$. 
Here, $\theta_e$ controls the spatial proximity sensitivity, while $\tau_e$ sets the minimum interaction ratio threshold for establishing entity relationships.
After identifying spatially adjacent entity pairs $\mathcal{R}$, we employ LLM GPT-4o~\cite{achiam2023gpt} to determine the semantic relationships between these entities: 
\begin{equation}
\label{eq:e_initial}
\mathcal{E}_\text{initial} = LLM(\{n_i, n_j, \vec{r}_{i,j}\}_{\{i,j\} \in \mathcal{R}}, \mathcal{P}_\text{edge}), 
\end{equation}
where $\vec{r}_{i,j}$ represents the 3D vector from the center of entity $i$ to the center of entity $j$, $\mathcal{P}_\text{edge}$ is the edge prompt (Fig.~\ref{fig:prompt}), and each node $n_i$ contains the spatial coordinates and semantic tag $\mathcal{O}_\text{tag}^i$ of entity $i$:
\begin{equation}
\begin{gathered}
n_i = \{(x_i, y_i, z_i), \mathcal{O}_\text{tag}^i\}, \\
\mathcal{O}_\text{tag} = LLM(\mathcal{O}_\text{caption}, \mathcal{P}_\text{tag}),
\end{gathered}
\end{equation}
where $(x_i, y_i, z_i)$ denotes the estimated center of entity $i$,
$\mathcal{O}_\text{tag}$ represents the entity tags derived from the original captions using the tag prompt $\mathcal{P}_\text{tag}$ (Fig.~\ref{fig:prompt}). 

Note that all entity pairs are processed simultaneously in a single LLM call to establish relationships, rather than individual pairwise inferences, which improves computational efficiency. Similarly, multiple entity captions are also processed through a single LLM inference to derive their corresponding tags.
While entity captions employ a fully open-set vocabulary to capture the diverse nature of object categories, the relationship prediction prioritizes selection from a set of common spatial relations (e.g., ``on'', ``under'', ``next to'') while retaining moderate open-set flexibility for uncommon cases. Unlike object categories which are inherently diverse and unlimited, spatial relationships constitute a relatively fixed and finite set of fundamental relations. Fully open-set relationship prediction would generate multiple varied expressions for identical spatial relations (e.g., ``on top of'', ``placed on'', ``resting on''), complicating query matching in robotic applications. This design balances expressiveness with computational efficiency for robotic task execution, where standardized relationship vocabularies enable faster query matching and more reliable task planning.

\subsubsection{Graph Update}
During the incremental mapping process, we continuously update the scene graph by incorporating new observations to enhance semantic understanding. The update process includes both refinement of existing nodes and integration of newly discovered entities, along with corresponding edge updates.

For node updates, we employ a confidence-based selective update strategy rather than fixed-interval updates of all nodes. The set of nodes requiring updates $\mathcal{N}_\text{update}$ includes: (i) existing nodes with confidence scores $\mathcal{C}_i$ (Eq.~\ref{eq19}) below the threshold $\tau_\text{update}$, and (ii) newly discovered entities from the mapping process detailed in Sec.~\ref{sec:Hierarchical 3D Semantic Optimization}. For selected nodes, we dynamically refine their multi-view representation $O_i$ (Eq.~\ref{eq:oi}). At time $t$, $O_i$ is updated as:
\begin{equation}
O_i^t = \{c_i', m_i^{1'}, m_i^{2'}, m_i^{3'}\},
\end{equation}
where $m_i^{j'}$ represent masked images from different views, $c_i'$ is the crop image, all selected based on geometric contribution. As new observations accumulate, we continuously update $O_i$ by prioritizing views with larger projected areas, which correspond to observations with better visibility and detail capture of the semantic entity. This geometric-guided selection enhances the quality of visual inputs for the VLM in subsequent node captioning.
Then, for each node in $\mathcal{N}_\text{update}$ with its updated multi-view representation $O_i^t$, we apply Eq.~\ref{eq:o_caption} to generate refined or new semantic captions, resulting in more accurate and detailed semantic descriptions. 
By selectively updating only uncertain nodes, this approach balances processing efficiency with scene graph accuracy.

For edge updates, we employ a similarly selective approach. Rather than recomputing all relationships, we only update edges connected to nodes that have been updated or added in the current iteration. For each such node $n_i \in \mathcal{N}_\text{update}$, we reevaluate its spatial adjacency with all other nodes $n_j \in \mathcal{N}$ in the scene graph using the same proximity analysis described in Eq.~\ref{eq:pij} and Eq.~\ref{eq:rij}. This process identifies candidate node pairs $\mathcal{R}_\text{update} = \{(i,j) \mid R_{ij} > \tau_e\}$ whose relationships require updating:
\begin{equation}
\mathcal{E}_\text{update} = LLM(\{n_i, n_j, \vec{r}_{i,j}\}_{\{i,j\}\in \mathcal{R}_\text{update}}, \mathcal{P}_\text{edge}).
\end{equation}
This targeted update strategy ensures both semantic accuracy and computational efficiency as the graph evolves. By selectively refining existing entities, incorporating new ones, and updating their interrelationships based on new observations, our approach enables progressive enhancement of the 3D scene understanding during exploration.

\begin{figure*}[t]
    \centering
    \includegraphics[width=\linewidth]{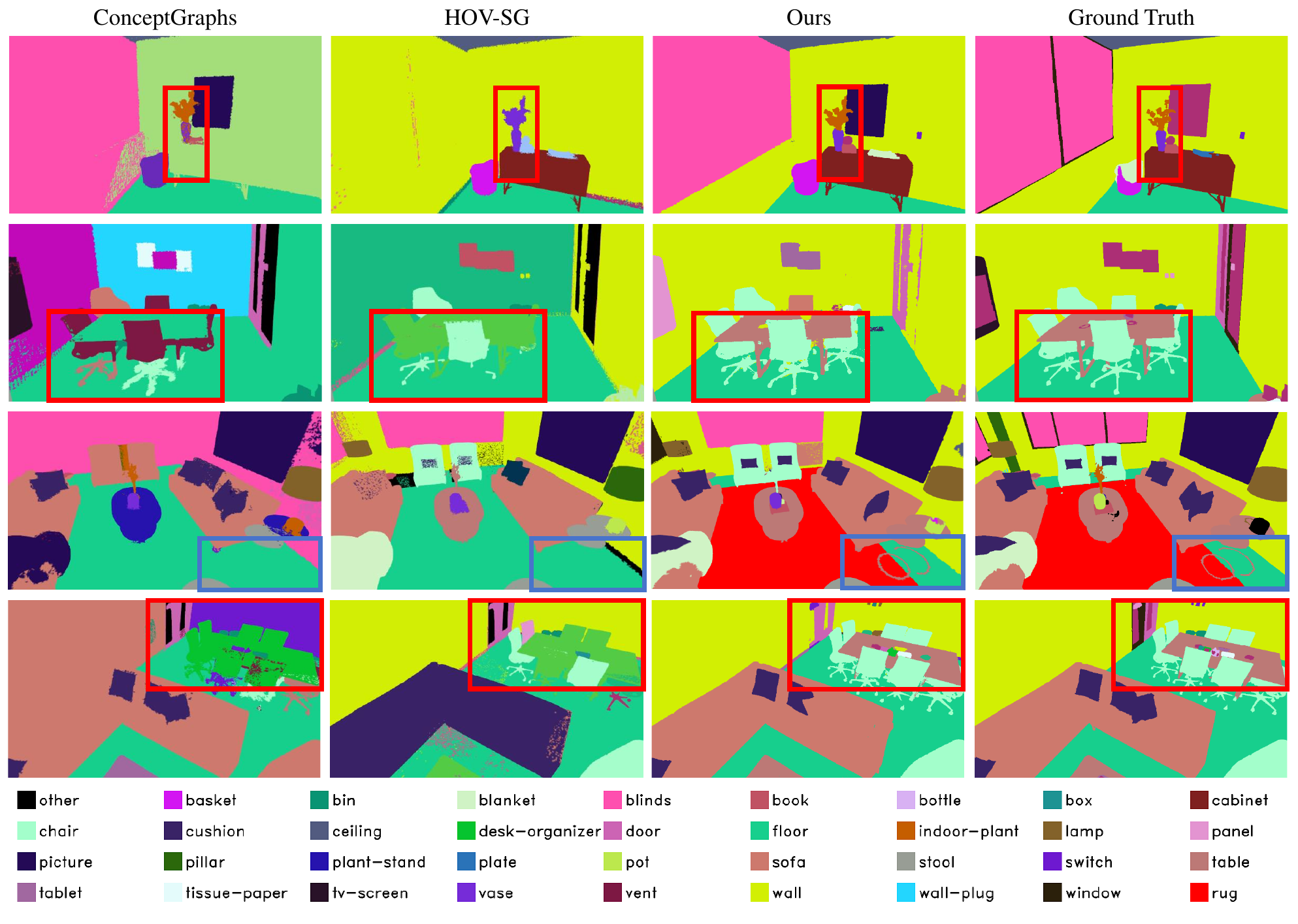}
    \caption{Qualitative comparison of semantic segmentation on Replica~\cite{straub2019replica}. Details are highlighted with colorful boxes. Our method achieves more precise semantic representation than baselines. 
    }
    \label{fig:vis-replica-2d}
\end{figure*}

\section{Experiments}
\subsection{Experimental Setup}
\subsubsection{Datasets}
We follow open-vocabulary semantic mapping methods~\cite{yang2025opengs, werby23hovsg, gu2024conceptgraphs} to conduct experiments on both Replica~\cite{straub2019replica} and ScanNet~\cite{dai2017scannet} datasets. These two datasets provide semantic ground truth annotations for semantic segmentation evaluation.
For Replica, we evaluate eight scenes: \textit{office0}-\textit{office4} and \textit{room0}-\textit{room2}. Regarding ScanNet, we evaluate five scenes: \textit{scene0011\_00}, \textit{scene0050\_00}, \textit{scene0231\_00}, \textit{scene0378\_00}, \textit{scene0518\_00}.
Moreover, we evaluate scene graph construction accuracy on 3RScan~\cite{Wald2019RIO}, as this dataset provides ground truth relationships. We select four representative scenes from 3RScan, including 137a8158, 10b17940, bf9a3dd3, 0988ea72. To further validate the practical applicability of our method, we conduct additional experiments on real-world scenes.

\subsubsection{Evaluation Metrics}
For semantic segmentation evaluation, we use \textit{mIoU (\%)}, \textit{F-mIoU (\%)}, and \textit{mAcc (\%)} metrics following \cite{werby23hovsg}.
For graph evaluation, we use \textit{Recall} metric for evaluation.

\subsubsection{Baselines}
We compare our method with open-vocabulary semantic mapping methods, including traditional representation-based~\cite{gu2024conceptgraphs, werby23hovsg} and 3DGS-based~\cite{zhou2024feature, ye2024gaussian, yang2025opengs} approaches.
Our approach leverages the high-fidelity rendering advantages of 3DGS. Therefore, for 2D novel-view semantic segmentation, we compare against 3DGS-based methods, as traditional representation-based approaches typically produce sparse 2D segmentation results with lower performance, making them less suitable benchmarks for this task. 
Regarding scene graph construction accuracy, we primarily compare with ConceptGraphs~\cite{gu2024conceptgraphs}, as the other aforementioned methods only implement 3D semantic segmentation without constructing relational graphs between objects.

\subsubsection{Implementation Details}
We employ fixed confidence thresholds throughout our system: high confidence threshold $\tau_\text{high} = 0.7$ and low confidence threshold $\tau_\text{low} = 0.3$. These thresholds are applied across all optimization stages to evaluate semantic reliability.
In Sec.~\ref{sec:Hierarchical 3D Semantic Optimization}, we set pixel overlap ratio threshold $\tau_m=0.7$, semantic ambiguity threshold $\tau_\text{valid} = 0.25$ and spatial-semantic coherence threshold $\tau_p = 0.15$ for local 3D optimization. The spatial search radius is set to $r_\text{search} = 0.1$. 
Moreover, we set threshold for viewpoint distinctiveness and geometric overlap as $\tau_\text{long}=0.5$ in Sec.~\ref{sec:Long-term global optimization}. Both global 3D refinement (Sec.~\ref{Global 3D refinement}) and long-term global optimization (Sec.~\ref{sec:Long-term global optimization}) are performed every 6 keyframes.
In Sec.~\ref{sec:Progressive Graph Construction}, we utilize spatial proximity threshold $\theta_e = 0.1$ and interaction ratio threshold $\tau_e = 0.2$ for entity relationship establishment.
Node updates are triggered when confidence falls below $\tau_\text{update} = \tau_\text{low} = 0.3$. Graph update is performed every 12 keyframes.
Additionally, we use Adam optimizer to optimize Gaussian parameters. We use a learning rate of 0.001 for scales optimization, 0.05 for opacity optimization, and 0.0001 for color optimization. The weighting coefficients of losses for gaussian optimization are $\lambda_{c}=1$ and $\lambda_{d}=0.1$.
We run OGScene3D on NVIDIA RTX 4090 GPU.

\begin{figure*}
    \centering
    \includegraphics[width=\linewidth]{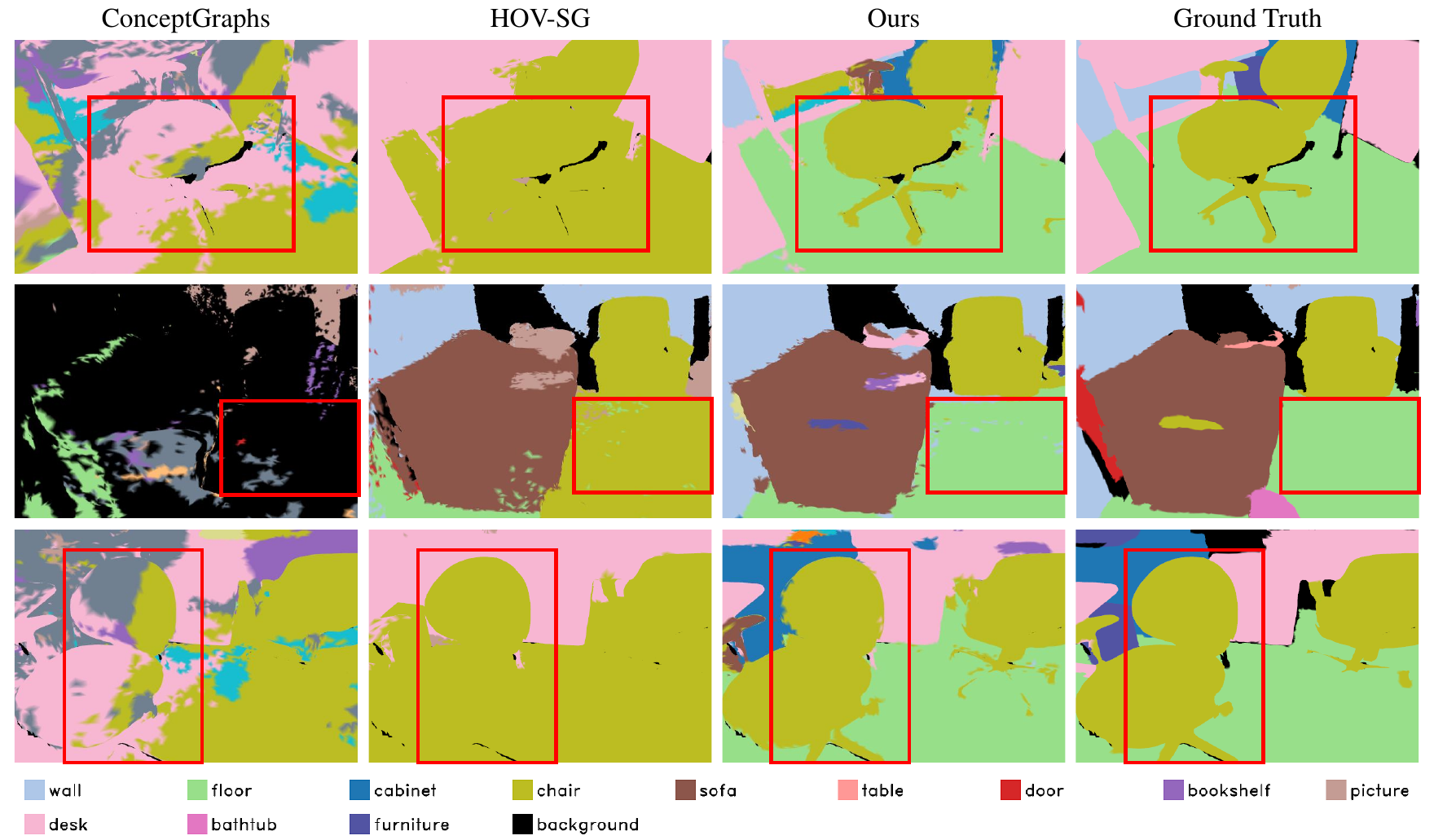}
    \caption{Qualitative comparison of semantic segmentation on real-world dataset ScanNet~\cite{dai2017scannet}. Details are highlighted with colorful boxes. Our method achieves more precise semantic mapping than baselines. 
    }
    \label{fig:vis-scannet}
\end{figure*}
\begin{table*}
\centering
    \caption{Comparison of zero-shot novel-view Semantic Segmentation.}
    \resizebox{\linewidth}{!}{
    \begin{tabular}{l|c| cccc cccc |c}
    \toprule
     Methods & Metrics & room0 & room1 & room2 & office0 & office1 & office2 & office3 & office4 & Avg. \\
    \midrule
     \multirow{2}{*}{Feature 3DGS~\cite{zhou2024feature}} & mIoU & 49.62 & 47.71 & 53.00 & 51.37 & 44.94 & 49.49 & 48.93 & 46.06 & 48.89 \\
      & mAcc & 59.46 & 55.37 & 62.21 & 59.42 & 52.19 & 58.25 & 57.84 & 55.36 & 57.51\\
    \midrule
     \multirow{2}{*}{GS-Grouping~\cite{ye2024gaussian}} & mIoU & 60.55 & 57.68 & 62.82 & 63.42 & 55.41 & 59.33 & 57.91 & 56.05 & 59.15 \\
      & mAcc & 69.02 & 67.83 & 71.41 & 73.61 & 66.78 & 71.51 & 70.29 & 69.06 & 69.94 \\
      \midrule
     \multirow{2}{*}{OpenGS-SLAM~\cite{yang2025opengs}} & mIoU & 64.06 & 61.18 & 64.38 & 61.53 & 58.11 & 63.05 & 62.71 & 60.25 & 61.91\\
      & mAcc & 73.53 & 72.13 & 74.17 & 73.05 & 69.03 & 75.36 & 73.62 & 74.01 & 73.11\\
      \midrule
     \multirow{2}{*}{OGScene3D (Ours)} & mIoU & \textbf{67.94} & \textbf{65.96} & \textbf{71.23} & \textbf{70.82} & \textbf{74.08} & \textbf{72.99} & \textbf{75.57} & \textbf{75.58} & \textbf{71.77}\\
      & mAcc & \textbf{91.33} & \textbf{87.71} & \textbf{83.60} & \textbf{92.73} & \textbf{92.88} & \textbf{83.08} & \textbf{90.85} & \textbf{91.04} & \textbf{89.15}\\ 
    \bottomrule
  \end{tabular}}
  \label{tab:replica2d}
\end{table*}

\subsection{Semantic Mapping Evaluation}
\label{sec:Semantic Mapping Evaluation}
We evaluate the accuracy of open-vocabulary semantic mapping through both 2D and 3D semantic segmentation. 
In all visualizations, different semantic categories are assigned distinct colors to clearly illustrate segmentation accuracy and highlight differences between methods.

\subsubsection{2D Semantic Segmentation}
To evaluate novel-view 2D semantic segmentation accuracy, we compute the \textit{mIoU} and \textit{mAcc} metrics between the rendered semantic maps and ground truth labels, following~\cite{yang2025opengs}. The quantitative results in Tab.~\ref{tab:replica2d} demonstrate that our method achieves superior novel-view segmentation performance across all test scenes, with an average 16\% improvement in \textit{mIoU} metric and 22\% improvement in \textit{mAcc} metric compared to baselines.

Fig.~\ref{fig:vis-replica-2d} shows 2D semantic segmentation comparisons on the Replica dataset~\cite{straub2019replica}. The baselines exhibit poor performance in both large background areas (walls, blinds) and fine-grained semantic details. For instance, baseline methods fail to correctly classify the flower as ``indoor plant'', instead erroneously merging it with the ``vase'' category. Similarly, they struggle with accurate segmentation of chair legs and table boundaries, demonstrating significant limitations in handling complex geometric structures and semantic boundaries. For the table leg, which is highlighted in blue, only our method successfully segments this detail, while other methods incorrectly group the table leg and rug with the floor.

Fig.~\ref{fig:vis-scannet} demonstrates 2D segmentation comparisons in real-world dataset ScanNet~\cite{dai2017scannet}. Real-world scenes present greater noise challenges, yet our method correctly distinguishes floors and chairs, while other methods incorrectly merge foreground objects with backgrounds, failing to achieve accurate segmentation results.

\begin{table*}[h]
  \centering
    \caption{Quantitative comparison of open-vocabulary 3D semantic segmentation on Replica dataset.}
  \resizebox{\linewidth}{!}{
  \begin{tabular}{l| l|cccc cccc|c}
    \toprule
    Methods & Metric& room0 & room1 & room2 & office0 & office1 & office2 & office3 & office4 & Avg.\\
    \midrule
    \multirow{3}{*}{ConceptGraphs~\cite{gu2024conceptgraphs}}
    & mIoU & 12.35 & 14.04 & 10.58 & 20.23 & 11.61 & 19.08 & 9.71 & 22.04 & 14.95\\
    & F-mIoU & 21.20 & 15.75 & 27.16 & 32.20 & 11.54 & 25.85 & 11.64 & 42.44 & 23.47\\
    & mAcc & 27.08 & 34.68 & 27.34 & 31.73 & 31.49 & 29.27 & 26.73 & 41.22 & 31.19\\
    \midrule
    \multirow{3}{*}{HOV-SG~\cite{werby23hovsg}}
    & mIoU & 33.74 & 27.37 & 18.27 & 24.81 & 11.93 & 20.98 & 19.95 & 27.37 & 23.05\\
    & F-mIoU & \textbf{62.77} & 40.84 & \textbf{43.78} & \textbf{33.78} & 17.78 & 40.51 & 26.00  & 43.58 & 38.63\\
    & mAcc & 43.57 & 39.22 & 26.18 & 28.17 & 23.88 & 24.99 & 28.46 & 28.72 & 30.40\\
    \midrule
    \multirow{3}{*}{OGScene3D (Ours)}
    & mIoU & \textbf{39.87} & \textbf{36.07} & \textbf{31.63} & \textbf{28.83} & \textbf{17.56} & \textbf{25.67} & \textbf{25.96} & \textbf{35.71} & \textbf{30.16}\\
    & F-mIoU & 53.23 & \textbf{42.22} & 40.68 & 28.02 & \textbf{22.78} & \textbf{54.32} & \textbf{51.32} & \textbf{54.69} & \textbf{43.41}\\
    & mAcc & \textbf{51.63} & \textbf{55.16} & \textbf{49.93} & \textbf{40.58} & \textbf{37.01} & \textbf{30.53} & \textbf{30.57} & \textbf{41.46} & \textbf{42.11}\\
    \bottomrule
  \end{tabular}}
  \label{tab:replica3d}
\end{table*}

\begin{table*}[h]
  \centering
    \caption{Quantitative comparison of open-vocabulary 3D semantic segmentation on ScanNet dataset.}
    \begin{threeparttable}
  \resizebox{\linewidth}{!}{
  \begin{tabular}{l| l|cccc c|c}
    \toprule
    Methods & Metric& scene0011\_00 & scene0050\_00 & scene0231\_00 & scene0378\_00 & scene0518\_00 & Avg.\\
    \midrule
    \multirow{3}{*}{ConceptGraphs~\cite{gu2024conceptgraphs}}
    & mIoU & -- & -- & -- & -- & -- & 16.00\\
    & F-mIoU & -- & -- & -- & -- & -- & 20.00 \\
    & mAcc & -- & -- & -- & -- & -- & 28.00\\
    \midrule
    \multirow{3}{*}{HOV-SG~\cite{werby23hovsg}}
    & mIoU & 27.80 & 17.23 & 22.72 & 26.67 & 16.58 & 22.20\\
    & F-mIoU & 42.19 & 36.50 & 26.39 & 29.15 & 17.27 & 30.30\\
    & mAcc & 44.59 & \textbf{41.62} & 40.50 & 51.28 & \textbf{37.51} & 43.10\\
    \midrule
    \multirow{3}{*}{OGScene3D (Ours)}
    & mIoU & \textbf{31.47} & \textbf{28.54} & \textbf{25.35} & \textbf{37.70} & \textbf{24.13} & \textbf{29.44}\\
    & F-mIoU & \textbf{42.33} & \textbf{50.53} & \textbf{26.93} & \textbf{38.61} & \textbf{23.29} & \textbf{36.34}\\
    & mAcc & \textbf{55.55} & 40.77 & \textbf{42.21} & \textbf{58.46} & 35.05 & \textbf{46.41}\\
    \bottomrule
  \end{tabular}}
  \\[3pt]
    \footnotesize
    \raggedright
    Results of other methods are directly taken from the paper of HOV-SG~\cite{werby23hovsg}. Since HOV-SG~\cite{werby23hovsg} only reports average accuracy metrics for ConceptGraphs~\cite{gu2024conceptgraphs} without per-scene breakdowns, the per-scene accuracy values for \cite{gu2024conceptgraphs} are marked with "--" in the table.
  \end{threeparttable}
  \label{tab:scannet3d}
\end{table*}
  
\begin{figure*}
    \centering
    \includegraphics[width=\linewidth]{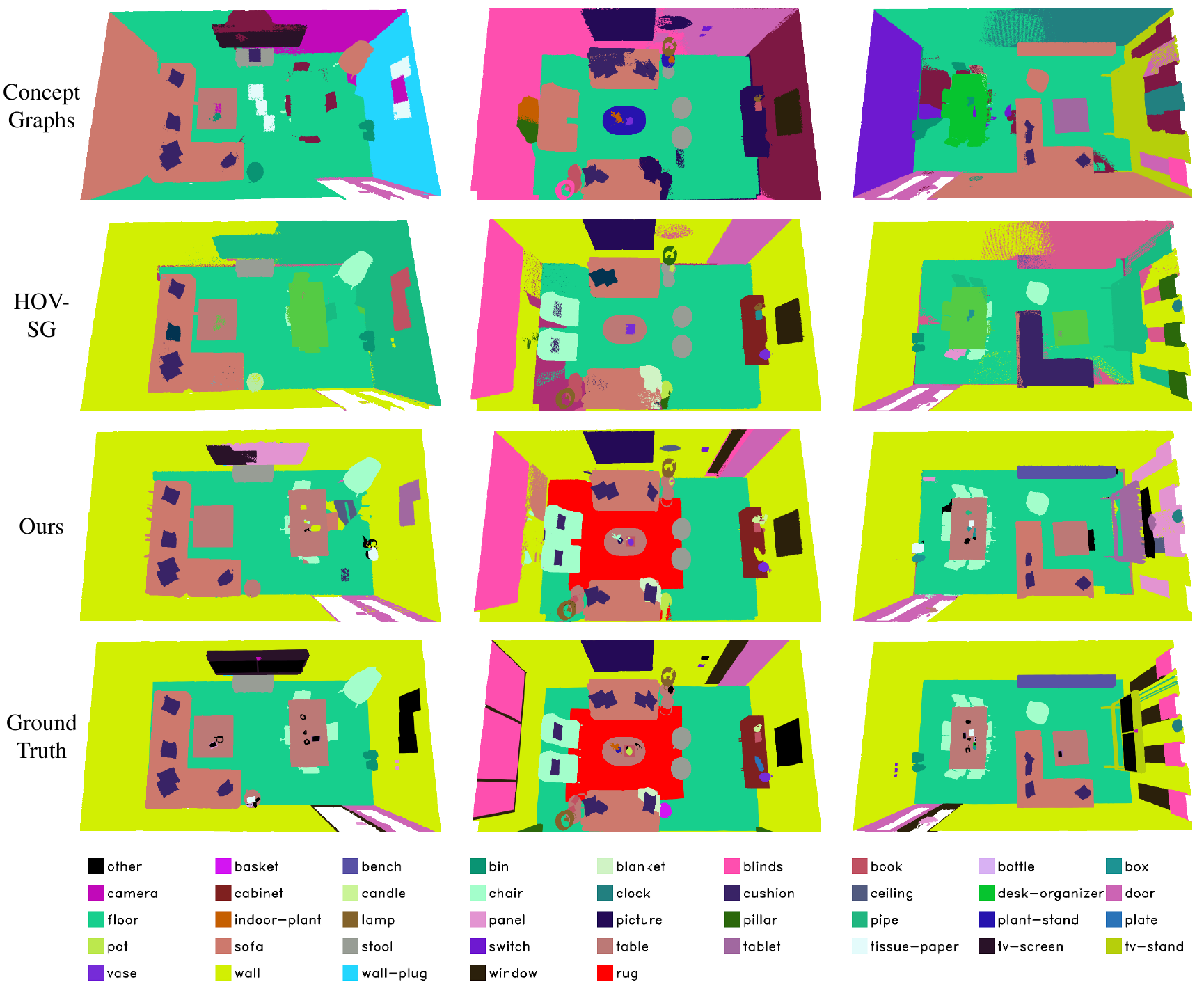}
    \caption{Qualitative comparison on 3D semantic mapping. Our method achieves predictive completion, enabling more accurate and comprehensive 3D semantic reconstruction results. Only our method can correctly segment objects such as \textcolor{myred}{\textbf{rug}}, \textcolor{mybrown}{\textbf{table}} and background wall.
    }
    \label{fig:vis-replica-3d}
\end{figure*}

\subsubsection{3D Semantic Segmentation}
We follow \cite{werby23hovsg} for 3D segmentation evaluation. While our method directly outputs semantic meanings rather than features, there may be cases where our generated descriptions from LLM convey the same semantic meaning as ground truth labels but with different phrasing. For a fair comparison, we employ Gemma~\cite{team2025gemma} to establish correspondence between our outputs and scene labels before evaluation. In contrast, baseline methods~\cite{werby23hovsg, gu2024conceptgraphs} establish label correspondence through similarity computation between features and scene text descriptions.
Additionally, for consistency with baseline methods, we utilize ground truth poses for 3D semantic mapping during 3D segmentation evaluation, although our approach is capable of estimating poses during mapping. 
As shown in Tab.~\ref{tab:replica3d} and Tab.~\ref{tab:scannet3d}, our method outperforms baselines in average precision, achieving up to 36\% improvement on the Replica dataset~\cite{straub2019replica} and up to 28\% improvement on the ScanNet dataset~\cite{dai2017scannet}. Fig.~\ref{fig:vis-replica-3d} presents qualitative comparisons on open-vocabulary 3D semantic mapping. Compared to baselines, our method correctly segments large background walls as well as rug with appearance similar to the floor. At the object level, we accurately classify tables that other methods incorrectly assign to different categories.

Note that both our method and the baselines utilize the same SAM segmentation results as input, thus all methods start with identical segmentation errors and multi-view inconsistencies. However, our method demonstrates superior performance in open-set semantic mapping compared with baselines. This improvement can be attributed to our confidence-based semantic representations, combined with 3D confidence optimization and long-term optimization. 
By employing hierarchical 3D optimization and leveraging historical information for refinement, our OGScene3D enables accurate segmentation and semantic labeling even when objects are incorrectly segmented in certain frames or partially occluded across different viewpoints.

\begin{table}
  \centering
    \caption{
      Quantitative comparison of scene graph for \textit{Recall(\%)} metric on 4 scenes of the 3RScan dataset.
      }
    \resizebox{\linewidth}{!}{
      \begin{tabular}{l|cccc}
        \toprule
        Methods & 137a8158 & 10b17940 & bf9a3dd3 & 0988ea72\\
         \midrule
         ConceptGraphs~\cite{gu2024conceptgraphs} & 2.5 & fail & 3.5 & 0.9\\
         OGScene3D (Ours) &\textbf{28.7} & \textbf{25.6}& \textbf{22.8}& \textbf{18.3}\\
        \bottomrule
      \end{tabular}}
  \label{tab:3rscan}
\end{table}

\subsection{Scene Graph Evaluation}
Since node accuracy has already been evaluated in the 3D semantic mapping section (Sec.~\ref{sec:Semantic Mapping Evaluation}), we primarily focus on evaluating the accuracy of relationships in the scene graph construction.
As shown in Tab.~\ref{tab:3rscan}, baseline struggles to accurately identify relationships in real-world scenes, while our approach achieves superior open-set relationship estimation results. Note that the original ConceptGraphs~\cite{gu2024conceptgraphs} was limited to recognizing only three types of relationships(``in'', ``on'', ``close by''), we extended its relationship vocabulary to cover the full range of relationships present in the 3RScan dataset~\cite{Wald2019RIO} for fair comparison.
Our method achieves superior performance due to our confidence-based updating strategy, which continuously integrates new observations to optimize the scene graph.  This approach enables more accurate edge estimation by refining relationship predictions as more viewpoints become available during exploration.

\begin{figure}
    \centering
    \includegraphics[width=\linewidth]{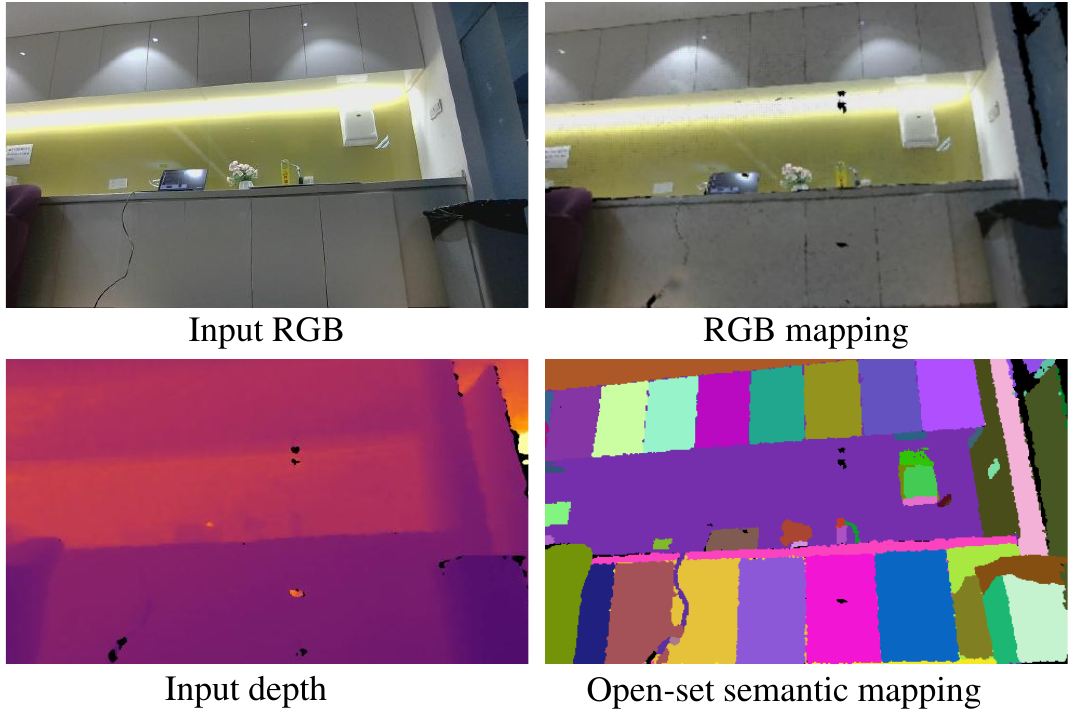}
    \caption{Visualization of real-world experiments. Using RGB-D data from the Gemini 336 camera as input, our method performs accurate dense RGB mapping and open-set semantic mapping.
    }
    \label{fig:real-world-vis}
\end{figure}

\begin{figure}
    \centering
    \includegraphics[width=\linewidth]{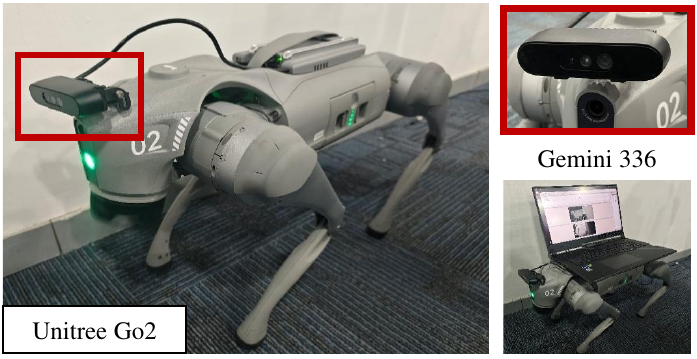}
    \caption{The hardware setup used for real-world semantic mapping and scene understanding experiments.
    }
    \label{fig:real-world-hardware}
\end{figure}

\subsection{Real World Experiment}
We conducted real-world experiments using a Unitree Go2 quadruped robot equipped with a Gemini 336 RGB-D camera, as shown in Fig.~\ref{fig:real-world-hardware}. All computations are performed on a laptop featuring an NVIDIA RTX 3080 GPU. The experiments involved the robot moving through indoor environments while simultaneously performing semantic mapping and scene understanding. Fig.~\ref{fig:real-world-vis} shows representative results from our real-world experiments, demonstrating that our method achieves high-quality RGB mapping and open-set semantic mapping.
Detailed experimental results are presented in the video demo.

\begin{table}
  \centering
    \caption{
      Runtime comparison on ScanNet dataset. (m) minutes (s) seconds.
      }
    \resizebox{\linewidth}{!}{
      \begin{tabular}{l|c|ccc}
        \toprule
        Methods & Time/scene & mIoU & F-mIoU & mAcc\\
         \midrule
         ConceptGraphs~\cite{gu2024conceptgraphs} & 122m5s & 16.00 & 20.00 & 28.00\\
         HOV-SG~\cite{werby23hovsg} & 31m56s & 22.20 & 30.30 & 43.10\\
         OGScene3D (Ours) & \textbf{10m13s} & \textbf{29.44} & \textbf{36.34} & \textbf{46.41}\\
        \bottomrule
      \end{tabular}}
  \label{tab:runtime}
\end{table}

\begin{table}[t]
  \centering
    \caption{
       Detailed runtime of our method regarding open-set semantic mapping.
      }
      \footnotesize
      \begin{tabular}{ccc|c}
        \toprule
         Local 3D & Global 3D & Long-term & Per-frame\\
         Optimization & Refinement & Optimization & Runtime \\
         (s) & (s) & (s) & (s) \\
         \midrule
         0.248 & 0.055 & 0.020 & 0.323\\
        \bottomrule
      \end{tabular}
  \label{tab:runtime-detailed}
\end{table}

\subsection{Runtime Analysis}
We analyze the runtime of our method, which simultaneously constructs semantic maps and scene graphs. Since some baseline methods only perform semantic mapping, we first compare the runtime for 3D semantic mapping tasks.
Both OGScene3D and baselines utilize SAM~\cite{kirillov2023segment} for 2D open-set segmentation before performing 3D mapping.  Given that SAM~\cite{kirillov2023segment} is computationally intensive and could be replaced with lighter segmentation networks in practical applications, we exclude the SAM runtime from our measurements. This exclusion provides a more direct comparison of the core mapping methods' efficiency.
As shown in Tab.~\ref{tab:runtime}, where runtime represents the processing duration for a complete scene, our method requires significantly less computational time while simultaneously achieving higher accuracy compared to baseline approaches, demonstrating the efficiency-accuracy balance of our approach.
Additionally, we report the runtime of each module in our semantic mapping, as shown in Tab.~\ref{tab:runtime-detailed}. Since these three modules operate at different execution frequencies, we calculate the total runtime and then average it per frame for detailed analysis. 

\begin{figure}[t]
    \centering
    \includegraphics[width=\linewidth]{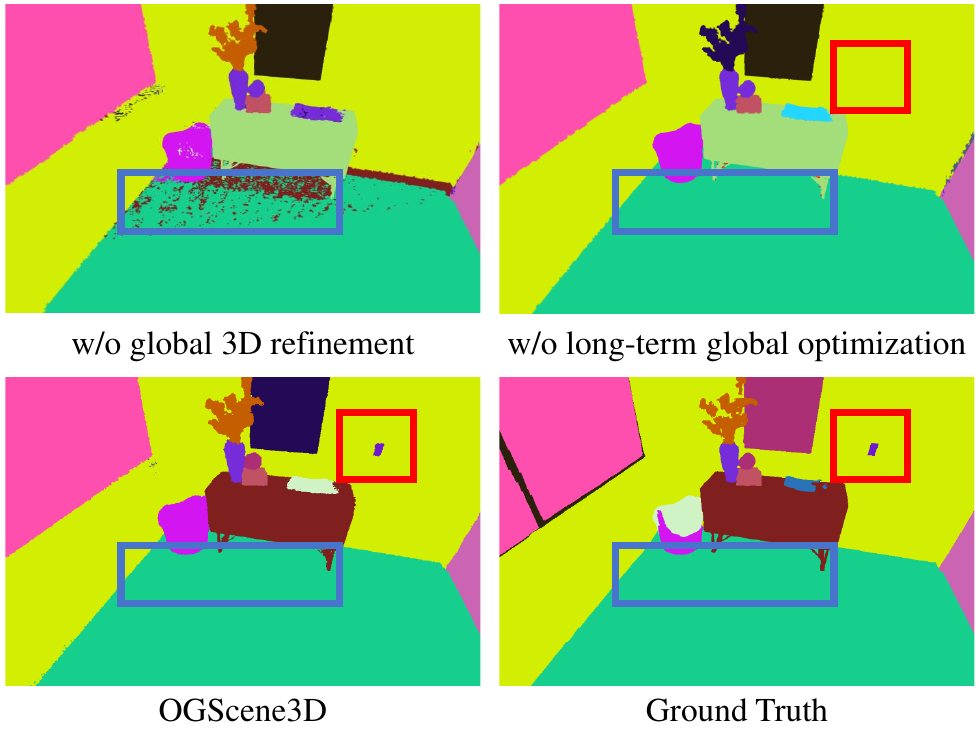}
    \caption{Visualization of ablation study. Semantic rendering results and ground truth labels are shown on Replica dataset~\cite{straub2019replica}. \textcolor{myblue}{Blue boxes} show degraded results without global 3D refinement. \textcolor{myred}{Red boxes} demonstrate successful segmentation of sparsely observed objects through long-term global optimization.
    }
    \label{fig:vis-ablation}
\end{figure}

\begin{table}[t]
  \centering
    \caption{
      Ablation study of our innovations on Replica dataset.
      }
    \resizebox{\linewidth}{!}{
      \begin{tabular}{l|l|ccc}
        \toprule
        & Methods & mIoU & F-mIoU & mAcc\\
         \midrule
         (a) & Ours (only Local 3D optimization) & 22.98 & 35.15 & 42.85\\
         (b) & Ours ($+$ Global 3D refinement to (a)) & 25.35 & 38.16 & 49.02\\
         (c) & Ours (full, $+$ Long-term global optimization to (b)) & \textbf{36.07} & \textbf{42.22} & \textbf{55.16} \\
        \bottomrule
      \end{tabular}}
  \label{tab:ablation}
\end{table}

\subsection{Ablation Study}
We conduct ablation studies of our proposed strategies and modules in this section.

\subsubsection{Effectiveness of Hierarchical 3D Semantic Optimization}
Our hierarchical 3D semantic optimization consists of two key components: local 3D optimization and global 3D refinement.
The local 3D optimization performs multi-frame label association, which is fundamental to the system's functionality. Therefore, we cannot completely remove the hierarchical optimization framework for evaluation.
To assess the effectiveness of this module, we conduct an ablation study by retaining only the basic local correspondence while removing the global 3D refinement component, as validated in Tab.~\ref{tab:ablation}(b).
The visual results are illustrated in the blue boxes of Fig.~\ref{fig:vis-ablation}. Without global 3D refinement, the reconstructed scenes exhibit noticeable artifacts, including noisy semantic points on the ground surface and erroneous semantic clusters. 
This degradation occurs because the global refinement module performs semantic optimization at both cluster-level and individual Gaussian corrections, effectively refining low-confidence semantic predictions from different perspectives to achieve high-precision semantic mapping.

\subsubsection{Effectiveness of Long-term Global Optimization}
Tab.~\ref{tab:ablation}(c) demonstrates that long-term global optimization significantly enhances the accuracy of 3D semantic mapping. As highlighted by the red boxes in Fig.~\ref{fig:vis-ablation}, the wall-mounted switch presents a challenging case where, despite being visible from multiple viewpoints, SAM fails to segment it accurately when observed from distant positions. This is a case of sparse observation where individual viewpoints provide insufficient detail for reliable segmentation. Our long-term optimization module addresses this limitation by integrating historical observations across multiple frames, enabling precise segmentation and modeling of sparsely observed entities that would otherwise be missed or inaccurately represented in the semantic map.
\section{Conclusion and future work}
\subsection{Conclusion}
We propose OGScene3D, an open-vocabulary scene understanding system that enables accurate 3D semantic mapping and scene graph construction incrementally. We introduce a confidence-based Gaussian semantic representation that jointly models semantic predictions and their reliability, enabling robust scene modeling. We propose hierarchical 3D semantic optimization that performs local correspondence establishment and global refinement, achieving globally consistent semantic maps throughout the incremental exploration process. We propose long-term global optimization that leverages long-term historical observations to continuously refine semantic understanding through 2D-3D semantic consistency and Gaussian rendering contribution. We develop a progressive graph construction approach that dynamically creates and updates both nodes and semantic relationships, enabling continuous optimization of 3D scene graphs during robotic exploration.
Our system has applications in autonomous robotic systems and augmented reality environments, where accurate 3D semantic understanding is essential for real-world deployment. The constructed semantic maps and scene graphs can be leveraged for high-level task planning, spatial reasoning, and contextual scene queries across diverse robotic and AR applications. Moreover, our research establishes a foundation for future advancements in autonomous decision-making and spatial reasoning, paving the way for more sophisticated task planning and environmental interaction capabilities in intelligent systems.

\subsection{Limitation and Future Work}
A main limitation of OGScene3D involves its capacity to model only spatial relationships, lacking exploration of complex inter-entity relationships such as functional associations. 
In future work, we aim to extend our system to achieve comprehensive multi-relational scene understanding for real-time robotic deployment.


%

\bibliographystyle{IEEEtran}
\bibliography{IEEEabrv,main}

\newpage




\vfill

\end{document}